\documentclass[oneside]{article}

\usepackage{PRIMEarxiv}

\usepackage{amsthm}

\newtheorem{definition}{Definition}

\usepackage[utf8]{inputenc} 
\usepackage[T1]{fontenc}    
\usepackage{amsmath}    
\usepackage{hyperref}       
\usepackage{url}            
\usepackage{booktabs}       
\usepackage{bbm}
\usepackage{amssymb}
\usepackage{nicefrac}       
\usepackage{microtype}      
\usepackage{xcolor}         
\usepackage{graphicx}
\usepackage{subcaption}
\usepackage{colortbl}
\usepackage{algorithm}
\usepackage{algorithmicx}%
\usepackage{algpseudocode}%
\usepackage{listings}
\usepackage{enumitem}
\usepackage{tikz}
\usepackage{siunitx}
\usetikzlibrary{3d}
\usetikzlibrary{calc}
\usepackage{tikz-3dplot}
\usepackage{comment}

\usepackage[square, sort, numbers]{natbib}
	\setcitestyle{square}

\usepackage[labelfont=bf,font=small]{caption}

\graphicspath{ {./figures/} }

\newcommand{\NEW}[1]{{\color{blue} #1}}

\usepackage[normalem]{ulem} 
\newcommand{\NOTE}[1]{{\color{red} #1}}

\newcommand{\Uniforce}{UniForCE\xspace}

\newcounter{marginNoteCounter}
\newcommand{\mN}[1]{\stepcounter{marginNoteCounter}\,{\footnotesize $^{\text{\color{red}\textbf\themarginNoteCounter}}$}\marginpar{\footnotesize $^\text{\themarginNoteCounter}$\,\NOTE{#1}}}

\newsavebox\CBox
\def\textBF#1{\sbox\CBox{#1}\resizebox{\wd\CBox}{\ht\CBox}{\textbf{#1}}}
\usepackage{dsfont}

\newcommand{\myvar}[1]{\ensuremath{\operatorname{#1}}}

\newcommand{\bigO}{\mathcal{O}} 
\renewcommand{\ldots}{...} 

\makeatletter
\newcommand*{\centerfloat}{%
  \parindent \z@
  \leftskip \z@ \@plus 1fil \@minus \textwidth
  \rightskip\leftskip
  \parfillskip \z@skip}
\makeatother

\usepackage{amssymb}
\usepackage{xspace}
\usepackage{wrapfig}

\title{UniForCE: The Unimodality Forest Method for Clustering and Estimation of the Number of Clusters}

\author{
Georgios Vardakas \\
Dept. of Computer Science and Engineering\\
University of Ioannina\\
GR 45110, Ioannina, Greece\\
\texttt{g.vardakas@uoi.gr}\\
\And
Argyris Kalogeratos\\
Centre Borelli, ENS Paris-Saclay\\
Université Paris-Saclay\\
91190 Gif-sur-Yvette, France\\
\texttt{argyris.kalogeratos@ens-paris-saclay.fr}\\
\And
Aristidis Likas \\
Dept. of Computer Science and Engineering\\
University of Ioannina\\
GR 45110, Ioannina, Greece\\
\texttt{arly@cs.uoi.gr} \\
}

\date{}

\begin{document}
\maketitle

\begin{abstract}
Estimating the number of clusters $k$ while clustering the data is a challenging task. An incorrect cluster assumption indicates that the number of clusters $k$ gets wrongly estimated. Consequently, the model fitting becomes less important. In this work, we focus on the concept of unimodality and propose a flexible cluster definition called \emph{locally unimodal} cluster. A locally unimodal cluster extends for as long as unimodality is locally preserved across pairs of  subclusters of the data. Then, we propose the \Uniforce method for locally unimodal clustering. The method starts with an initial overclustering of the data and relies on the \emph{unimodality graph} that connects subclusters forming unimodal pairs. Such pairs are identified using an appropriate statistical test. \Uniforce identifies maximal locally unimodal clusters by computing a spanning forest in the unimodality graph. Experimental results on both real and synthetic datasets illustrate that the proposed methodology is particularly flexible and robust in discovering regular and highly complex cluster shapes. Most importantly, it automatically provides an adequate estimation of the number of clusters.

\keywords{clustering \and unimodality \and number of clusters \and overclustering \and cluster aggregation \and dip-test.}
\end{abstract}

\section{Introduction}\label{sec:intro}

%
Clustering is a fundamental unsupervised machine learning problem, aiming to uncover the structure of a dataset. As it requires minimal knowledge about the data, clustering is used in numerous data-driven analysis tasks. Clustering is defined as the process of partitioning a set of objects into groups, the clusters, such that data within the same group share common characteristics and differ from data in other groups. Despite its simple definition, the lack of labeled examples makes clustering an ill-posed problem, as there are many different `reasonable' objectives to optimize with~\cite{jain1999data,jain2010data}.



An essential question to think about before clustering a dataset is the following: what is a meaningful cluster and how can it be represented? The first part of the question concerns the \emph{cluster assumption}, that is, the characteristics that a subset of data should exhibit in order to be considered a cluster. The second part, which is intertwined with the former, is how to represent mathematically a cluster of the assumed nature. \emph{Model-based} assumptions consider a probabilistic model for each cluster, e.g.~Gaussian~\cite{EM-Chapter-2006} or linear models~\cite{linear-cl-1990}. \emph{Prototype-based} assumptions partition data around objects that can be centroids~\cite{k-means-history-2007}, medoids~\cite{fast-k-medoids-2018}, synthetic prototypes~\cite{kalogeratos2011document}, or various other exemplars~\cite{affinity-prop-2007,exemplar-based-2007,density-peaks-2014}. There are also several \emph{density-based} assumptions, the most typical of which is the \emph{density-level-based} that relies on several thresholds expressing the minimum local density level that a continuous region should have in order qualify as a cluster~\cite{dbscan-1996,dbscan++2019}. \emph{Mode-seeking} approaches, on the other hand, use parametric density estimation to locate the area in a region where the density is maximized~\cite{carreira2015review,density-peaks-2014}.

This work relates to a different type of density-based cluster assumption that focuses on the \emph{density shape}. In most of the previous approaches, there are density shape assumptions that are either implicit (e.g.~prototype-based assumptions lead to convex-shaped clusters) or consequential, but not preconditions (e.g.~Gaussian mixture modeling would always fit with Gaussian components regardless of the validity of this assumption). The previously proposed explicit density shape cluster assumptions mainly concern Gaussianity~\cite{x-means-2000,g-means-2003,pg-means-2006}. Furthermore, the work in~\cite{dip-means-2012} was one of the first to introduce \emph{unimodality} as an explicit assumption for clustering multivariate data. Unimodality was assessed by the proposed \emph{dip-dist} criterion, a statistical methodology for unimodality testing of multivariate data that relies on multiple univariate unimodality tests (dip-tests)~\cite{dip-statistic-1985} performed on the distribution of pairwise distances between datapoints. 



Once a clustering model has been selected, a clustering solution of the assumed properties is usually produced by an algorithm that optimizes an appropriate objective function. Algorithmic approaches include: flat $k$-means~\cite{lloyd1982least} or expectation-maximization~\cite{dempster1977maximum}, agglomerative methods, incremental (divisive) algorithms that add clusters one by one, region growing or merging procedures, as well as hybrid approaches~\cite{jain2010data, Bishop2006PatternRA}. Among them, the \emph{incremental density-shape-based} approaches have two notable advantages: i) they are robust because they employ well-founded statistical tests, and ii) they offer a straightforward way to estimate automatically the number of the clusters $k$~\cite{x-means-2000,g-means-2003,pg-means-2006,dip-means-2012,pdip-means-2018}.

Determining the number of clusters $k$ during the optimization procedure is one of the most challenging problems in the field, 
especially in high-dimensions~\cite{adolfsson2019cluster}. Most methods require $k$ as input. Others claim to estimate $k$, but they essentially translate the problem into another, hopefully easier, problem, i.e. the tuning of their hyperparameters 
(e.g.~\cite{dbscan-1996,affinity-prop-2007,density-peaks-2014,carreira2015review,li2023multi}). There are also well-known protocols and heuristics for estimating the number of clusters by comparing clustering solutions for different values of $k$, e.g.~stability-based evaluation~\cite{stability-based-2004}, silhouette coefficient~\cite{rousseeuw1987silhouettes}, gap statistic, and others~\cite{JSSv061i06}.


Important to be noted, that an inadequate cluster assumption for a dataset makes more possible that $k$ gets also wrongly estimated, and also that the clustering result will be less informative. The existence of irregular cluster shapes is what makes most cluster assumptions to fail. Typical approaches for dealing with this issue depart for the original dataspace. Spectral embeddings~\cite{Luxburg2007ATO}, deep data embeddings~\cite{shaham2018spectralnet,leiber2021dip}, or data-dependent distance metrics, e.g.~diffusion maps~\cite{diffusionMaps2005}
 or path-based metrics~\cite{path-based-spectral2020}, they all aim to embed the data into a new vector space, where
-hopefully- the clusters would be nicely-shaped and/or 
far from each other in order to be recovered by typical methods. It should be stressed that those lines of work overlook the discussion about cluster assumptions, while estimating $k$ is usually beyond their scope. 

In this work, we study clustering in the original dataspace, aiming at developing a clustering methodology that is flexible enough to identify regular (e.g. a cloud of points) and irregular cluster shapes. One of the approaches that has been followed is to consider multi-prototypes as cluster representatives, and define variations of multiple-means clustering~\cite{multikmeans2019,li2023multi}. Another recipe is to first employ \emph{overclustering} to find small highly homogeneous \emph{subclusters}, and then try to combine them into larger and more complex cluster shapes, for instance via density-based~\cite{mehmood2017clustering}, graph-based~\cite{gao2021git,skeleton2022}, or other visualization-based schemes~\cite{Merging-kmeans2018}. Agglomerating subclusters is a long-known approach, and one of the initial propositions was to use it for reducing the sensitivity and complexity of hierarchical clustering~\cite{Zhao2005HierarchicalCA}, but it can also be useful for discovering irregular-shaped clusters through a proper cluster linkage criterion. To some limited extent, this has also been explored using the unimodality criterion~\cite{unimerging2014,region-unimerging-2018}, but with a simple merging methodology that prevents the identification of irregular-shaped clusters. 
%

\begin{figure}[t]
	\centering
    \begin{subfigure}{0.45\textwidth}
        \centering
        \includegraphics[scale=0.25]{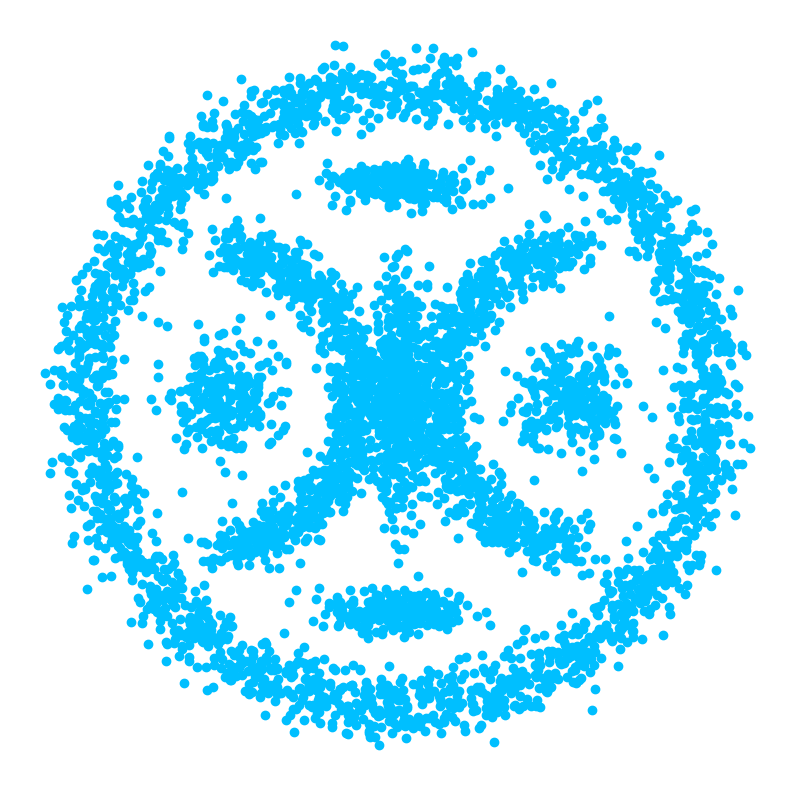}
        \subcaption{Synthetic dataset.}
        \label{fig:complex_data}
    \end{subfigure}
    \hspace{1.1em}
    \begin{subfigure}{0.45\textwidth}
        \centering
        \includegraphics[scale=0.25]{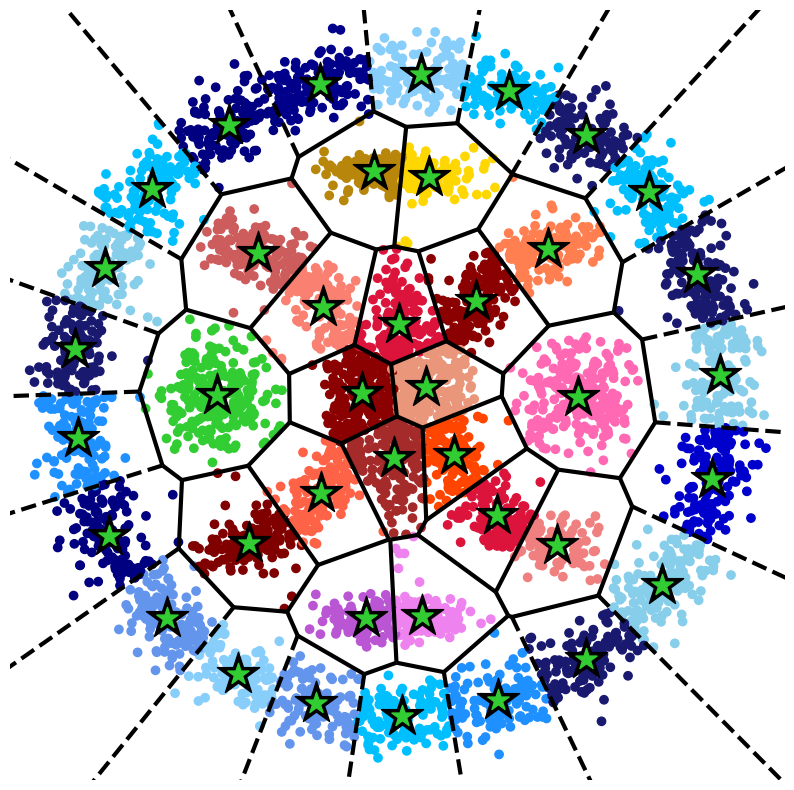}
        \subcaption{Convex overclustering.}
		\label{fig:K-plot-overclustering}
	\end{subfigure}
	\hspace{1.1em}
    \begin{subfigure}{0.45\textwidth}
        \centering
        \includegraphics[scale=0.25]{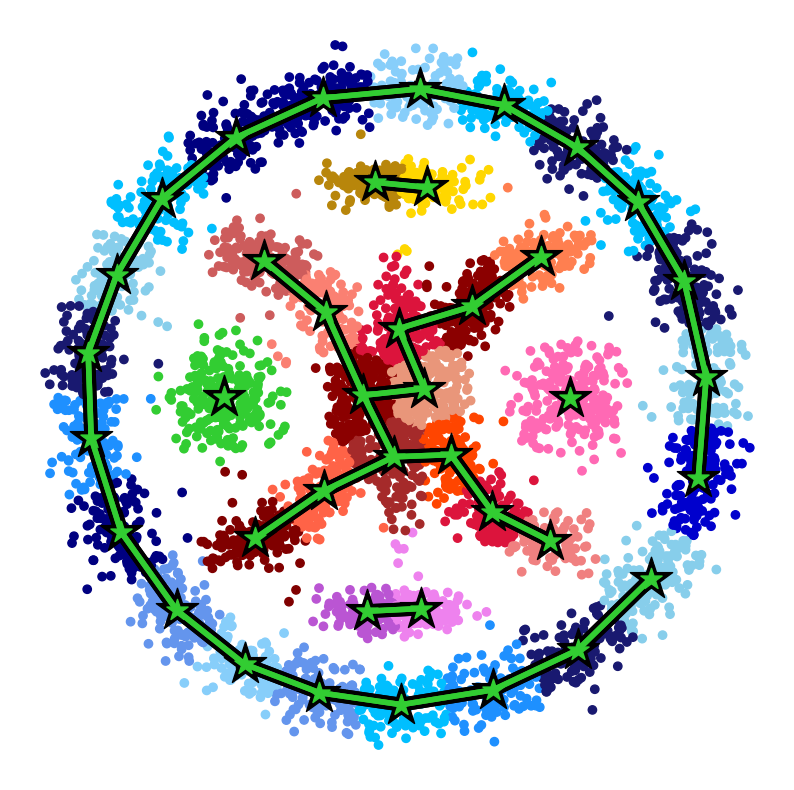}
        \subcaption{Spanning forest.}
		\label{fig:K-plot-mst}
    \end{subfigure}
    \hspace{1.1em}
    \begin{subfigure}{0.45\textwidth}
        \centering
        \includegraphics[scale=0.25]{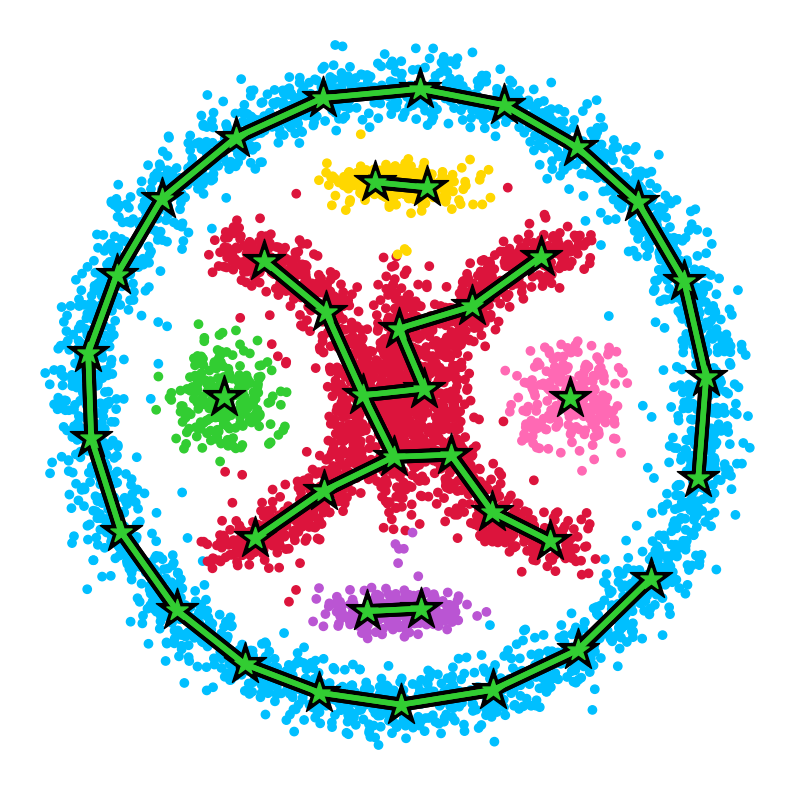}
        \subcaption{Locally unimodal clustering.}
		\label{fig:K-plot-spunimodal}
    \end{subfigure}
    \caption{Demonstration of the steps followed by the proposed \Uniforce clustering methodology on a synthetic dataset (Complex 2D). The input dataset is first overclustered into a large number of homogeneous subclusters lying in convex regions of the original dataspace. Then, based on pairs of subclusters that are jointly unimodal (unimodal pairs), a minimum spanning forest is computed, which provides the locally unimodal clusters as disconnected components.}
    \label{fig:K-plot}
\end{figure}

In this work, we focus on the concept of unimodality and propose a flexible cluster definition called \emph{locally unimodal cluster}. Such a cluster can be obtained by aggregating subclusters provided by an initial overclustering partition through a merging procedure that extends for as long as unimodality is locally preserved across pairs of subclusters. In order to examine this local property, we propose an effective statistical approach called \emph{unimodal pair testing} that relies on the univariate dip-test for unimodality~\cite{Hartigan1985}. We exploit these ideas and methods to propose a cluster aggregation approach, the \emph{Unimodality Forest for Clustering Estimation} method (\Uniforce), that boils down to: first, overclustering the dataset into small homogeneous subclusters lying in convex subregions, and then computing a \emph{spanning forest} over the \emph{unimodality graph} formed by the unimodal pairs of subclusters. Each spanning tree of the forest connects subclusters of the dataset that should be aggregated in the same final cluster, and the number of trees is an estimate for the number of clusters. Therefore, a maximal locally unimodal cluster extends for as long as unimodality is locally preserved. This feature makes our definition flexible enough to model typical unimodal as well as irregular-shaped clusters. An illustration of the main steps of our method is shown in Fig.\,\ref{fig:K-plot} using a complex synthetic dataset. Our extensive experimental study provide clear evidence that locally unimodal clusters model sufficiently real and synthetic datasets, and that our algorithmic design allows the robust estimation of the number of clusters while effectively partitioning the data. 

The rest of the paper is organized as follows. First, in Sec.\,\ref{sec:preliminaries}, we give some background elements and a brief literature background for graphs, overclustering and unimodality-based clustering methods. Then, in Sec.\,\ref{sec:LocallyUnimodalCluster}, we define the locally unimodal cluster. In Sec.\,\ref{sec:method} we present the  \Uniforce clustering method and provide its computational complexity. Finally, in Sec.\,\ref{sec:experiments}, we provide extensive experimental results and comparisons to real and synthetic datasets, while in Sec.\,\ref{sec:conclusion}, we present our conclusions and suggestions for future work.

\section{Preliminaries} 
\label{sec:preliminaries}

\noindent\textbf{Graph-related definitions.}~%
%
%
A \emph{weighted undirected graph} $G=(V,E)$ is endowed with a set of \emph{vertices} $V$ and a set of \emph{edges} $E$. Each edge connects an unordered pair of vertices, and the associated \emph{edge weight} represents the strength of this connection. The \emph{graph weight} is defined as the sum of all edge weights. %
%
%
A \emph{subgraph} of $G' \subseteq G$ is a graph between a subset of vertices $V' \subseteq V$ and a subset of edges $E' \subseteq E$ connecting them.
%
%
A \emph{path} between two vertices $V_i,V_j \in V$ is a subgraph of $G$ defined by a sequence of edges connecting distinct vertices, starting from $V_i$ and ending at $V_j$. Therefore, a path has no \emph{cycles}, as there is no repetition of vertices (consequently, same for edges), in the sequence. We say that $G$ is \emph{connected} if there exists a path between every pair of its vertices. The \emph{connected components} of $G$ are its maximal connected subgraphs.


A \emph{tree} $T=(V,E)$ is a connected acyclic graph, hence it holds $|E|=|V|-1$. A \emph{forest} $F$ is an acyclic graph composed by a collection of (possibly disconnected) trees. %
%
%
Any tree subgraph $T$ spanning over all the vertices of $G$, is called \emph{spanning tree}. A \emph{minimum spanning tree} (MST), though, is an spanning tree with minimal graph weight, while a graph may have more than one MST. Respectively, we can define the \emph{spanning forest}, or respectively the \emph{minimum spanning forest} (MSF), of a graph $G$. Given $G=(V,E)$, an MSF can be computed in $O(|E|\log{|V|})$ time by executing the classical Kruskal's algorithm~\cite{Kruskal1956}. As we will see next, the MSF is a mathematical object that is central in the way our methodology discovers and represents data clusters. 


%

\noindent\textbf{Clustering and overclustering.}~%
We are given a dataset $X$ containing $N$ datapoints in $\mathbb{R}^d$. Let $|\cdot|$ be the cardinal number of a set, $d$ the number of dimensions, and $i \in [N]$ imply $i \in \{1,...,N\}$. A \emph{partition} of  $X$ is a collection of pairwise-disjoint subsets whose union is exactly $X$. We are particularly interested in finding a \emph{cluster partition}  
of $X$ into a number of \emph{clusters} that exhibit additional intra- and/or inter-cluster structural properties. Clustering aims at revealing a true (but unknown) underlying partition $\mathcal{C}^*$, called ground-truth partition, and $k^*$ is the true (but unknown) number of clusters in $\mathcal{C}^*$. Most approaches solve this problem for several values of $k$ and select the best solution using appropriate criteria. This work is among the relatively few that perform data clustering while at the same time estimating the number of clusters $k^*$. 

Our approach relies on an initial overclustering of the dataset. We say that a clustering $\mathcal{C}^{+}$ of $X$ into $K>k^{*}$ clusters is an \emph{overclustering} of $X$, if it partitions the clusters of $\mathcal{C^*}$ into smaller subclusters, such that each subcluster is a subset of a cluster in $\mathcal{C^*}$. Fig.\,\ref{fig:K-plot-overclustering} presents an overclustering of the dataset into several subclusters. We denote a subclustering partition as $\mathcal{C}^{+} = \{c_1,...,c_K\}$, and the final partition containing larger clusters as $\mathcal{C} = \{C_1,...,C_k\}$, with $K>k$.


\noindent\textbf{Unimodality.}~%
The notion of \emph{unimodality} is a statistical property characterizing a probability density function. A univariate probability density $f$ is \emph{unimodal} if there exists a point $m\in\mathbb{R}$, called \emph{mode}, such that $f$ is non-decreasing in $(-\infty,m)$ and non-increasing in $(m,\infty)$.
Qualitatively, this means that $f$ admits its maximum value at the mode $m$ and may only stay the same or decrease as we move away from $m$. Contrary, two modes would emerge if there is a \emph{density gap}, i.e.~a considerable drop of the density level among two regions of higher density. Unimodality is a very generic property that allows for great density shape diversity. The densities of the distributions of many prominent families, such as the Uniform, Gaussian, Gamma and Beta, have this property. For the multivariate case, several definitions of unimodality have been proposed in the literature which are, however, not equivalent \cite{Dai_1989}. For further reading, we refer to the book by Dharmadhikari and Joag-Dev \cite{dharmadhikari1988unimodality}.

To test a \emph{univariate} dataset for unimodality, relevant statistical tests can be employed, such as Silverman's \emph{bandwidth test} \cite{Silverman1981,Hall2001} and the most prominent Hartigans' \emph{dip-test} \cite{Hartigan1985,Cheng1998}. The unimodality tests return a \emph{$p$-value} that is compared against a user-specified \emph{significance level} $\alpha$. For the \emph{multivariate} case, several definitions of unimodality have been proposed in the literature, for example \emph{star-unimodality} and \emph{generalized Anderson-unimodality} 
\cite{Dai_1989, dharmadhikari1988unimodality}.

Although there exist several mathematical definitions of multidimensional unimodality, testing whether a multidimensional dataset is unimodal is complicated and very few approaches have been proposed. The main idea of these approaches is to consider univariate representations of the data, and then test with univariate unimodality tests. \emph{Dip-dist}~\cite{dip-means-2012} proposed taking datapoints as `viewers' and aggregating multiple tests, each of them on the distribution of distances between a viewer and the rest of the data. The projected dip-means \cite{pdip-means-2018} applies the dip-test on univariate projections of the data on PCA and random axes. A different approach is the \emph{folding test}~\cite{folding-test-2018}, that relies on the empirical assumption that folding a multimodal distribution leads to a reduction in variance. Such an assumption is not always valid~\cite{chasani2022uu}.

\noindent\textbf{Unimodality-based clustering.}~%
By exploiting the previously mentioned criteria for assessing the unimodality of a multidimensional dataset, top-down (divisive) clustering methods have been proposed to partition a dataset into clusters that are unimodal according to the selected criterion. Those methods start with the whole dataset as a single cluster, and at each step they test each current cluster for unimodality. If the cluster is found multimodal, then it is split into two subclusters. The procedure is repeated until all clusters are considered as unimodal. The dip-means~\cite{dip-means-2012} and the projected dip-means~\cite{pdip-means-2018} algorithms  mentioned previously follow this strategy to perform clustering and simultaneously estimate the number of clusters. It should be emphasized that those methods cannot identify clusters of arbitrary shape, as it happens with the method we propose, which operates in a bottom-up fashion and assesses unimodality at a local (subcluster) level and not at the cluster level.

\section{Locally unimodal clusters}
\label{sec:LocallyUnimodalCluster}

\begin{figure}[t]
\centering
\begin{subfigure}{0.3\textwidth}
\centering
\includegraphics[scale=0.24]{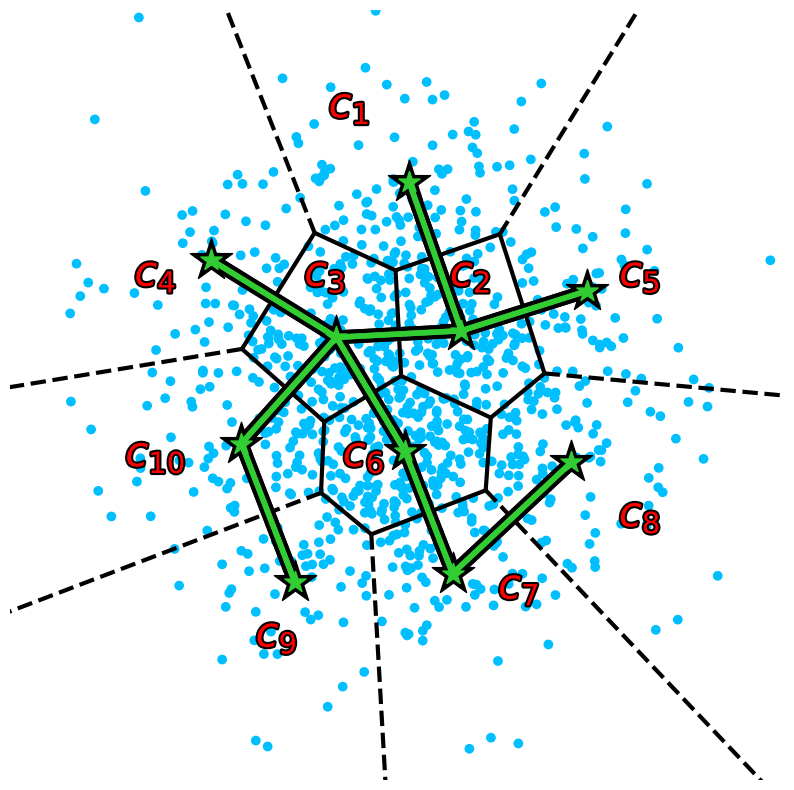}
\label{fig:LUC_Gaussian}
\subcaption{}
\end{subfigure}
\hspace{2mm}
\begin{subfigure}{0.3\textwidth}
\centering
\includegraphics[scale=0.24]{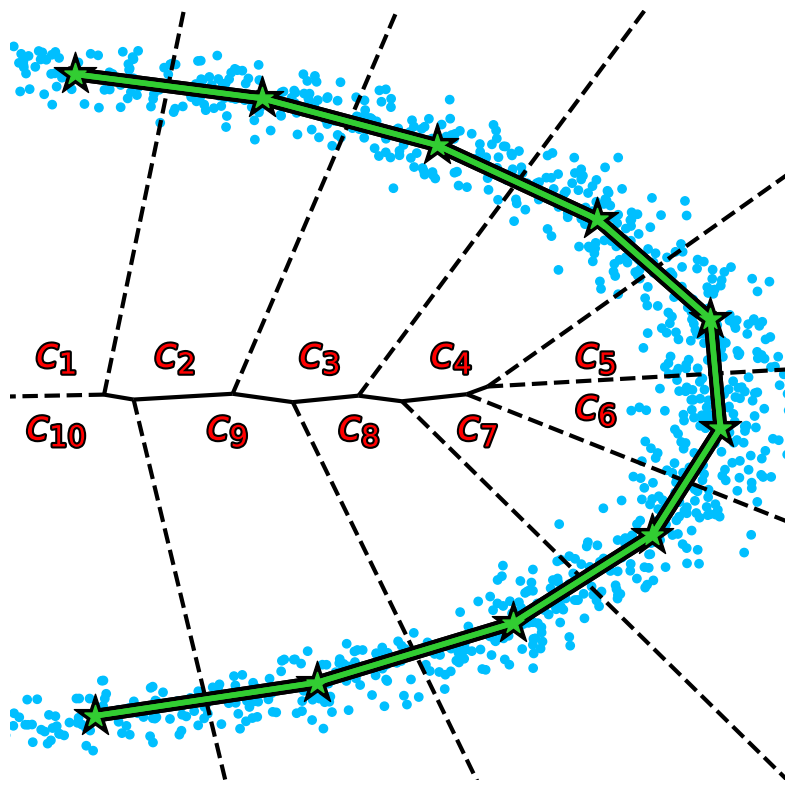}
\subcaption{}
\label{fig:LUC_Arc}
\end{subfigure}
\hspace{2mm}
\begin{subfigure}{0.3\textwidth}
\centering
\includegraphics[scale=0.24]{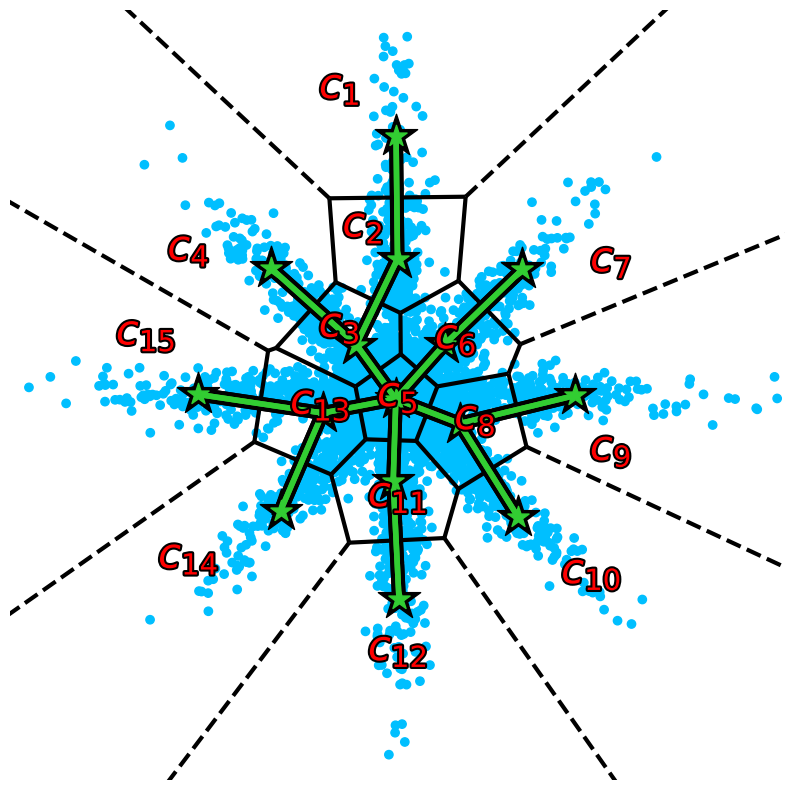}
\subcaption{}
\label{fig:LUC_Star}
\end{subfigure}
\caption{Examples of locally unimodal clusters. \textbf{a)}~A spherical Gaussian density. \textbf{b)}~An arc-shaped uniform density. \textbf{c)}~A star-shaped density composed by $3$ co-centric Gaussian ellipses. In each case, the data density is overclustered in subclusters, and a connected graph identifies the locally unimodal cluster by connecting with edges pairs of subclusters whose union is unimodal. Here, a spanning tree connects all the $K$ subclusters with the minimal number of $K-1$ edges. Such a tree, though, is only a subgraph of the unimodality graph, which may contain additional edges between subclusters that are unimodal pairs.
}
\label{fig:LUC}
\end{figure}
Our aim is to develop a clustering methodology that i) is able to capture complicated cluster structures, ii) can automatically discover the number of clusters, while at the same time iii) does not include any hard to tune hyperparameters. To this end, we introduce the concept of a \emph{locally unimodal cluster} by formulating how unimodality extends across neighboring subregions that are part of the same larger structure. We define the locally unimodal cluster as follows: 

\begin{definition}\label{def:LU-cluster}
\textbf{Locally unimodal cluster.}~A data subset $C\subseteq X$ is a locally unimodal cluster, if there is a partition $\mathcal{C}^{+}=\{c_1, \ldots, c_K\}$ of $C$ into subclusters 
lying in convex subregions, such that for every pair $(c_{i},c_{j})$ there exists a sequence $S_{ij}$ of 
 distinct subclusters, $S_{ij} = \{s_1=c_i, s_1,\ldots,s_{n-1}, s_n=c_j\}$, where the union of any two successive subclusters $s_{i} \cup s_{i+1}$ is unimodal.%
\end{definition}



A clustering partition $\mathcal{C}$ of $X$ is a \emph{locally unimodal clustering} if every cluster of $\mathcal{C}$ is locally unimodal. 

%
The locally unimodal cluster definition is flexible enough to encompass not only typical unimodal clusters, but also arbitrary-shaped clusters. Fig.\,\ref{fig:LUC} presents illustrative examples of locally unimodal clusters, where each $c_i$ is a subcluster and an edge between subclusters indicates that their union is unimodal. 

This cluster definition can be exploited in a bottom-up clustering framework that would start from an overclustering into a sufficient number of homogeneous subclusters $c_i$ lying in convex subregions, which can be computed by a typical partitional algorithm such as the $k$-means. Then, a way to identify \emph{unimodal pairs}, i.e.~subcluster pairs whose union is unimodal, need to be defined. Two subclusters forming a unimodal pair are expected to lie close to each other since the union of distant subclusters typically demonstrates a density gap that objects unimodality. The existence of unimodal pairs enables the union of small homogeneous subclusters to larger locally unimodal clusters, and this can be accomplished in a statistically sound manner. Our technique for deciding if two subclusters form a unimodal pair is presented in the next section. 

Once the initial overclustering partition is computed, we can define the corresponding \emph{unimodality graph} having the subclusters as vertices and an edge between each unimodal pair of subclusters. Note that a path in the unimodality graph defines a sequence of subclusters such that successive subclusters in the sequence form unimodal pairs. We call such a path as \emph{unimodal path}. Based on the above description, it is evident that the union of subclusters corresponding to any connected subgraph of the unimodality graph provides a locally unimodal cluster of arbitrary shape.
This is due to the fact that, since the subgraph is connected, there exists such a unimodal path between any two subclusters. The connected components of the unimodality graph correspond to maximal locally unimodal clusters
and define the clustering solution that provided by our method. The details of our method are described next. 

\section{Clustering methodology and the \Uniforce algorithm}
\label{sec:method}
In this section, we present the proposed clustering methodology and an algorithm implementing it, called \emph{Unimodality Forest for Clustering Estimation} (\Uniforce). The methodology determines the maximal locally unimodal clusters by finding the connected components of the unimodality graph, as explained in the section Sec.\,\ref{sec:LocallyUnimodalCluster}. 
The general methodological framework is given in Alg.\,\ref{alg:general}, and it comprises three main modules: \emph{Overclustering}, \emph{Unimodal pair testing}, and finally \emph{Clustering by subcluster aggregation}. 
In the subsections that follow, we present in detail how \Uniforce implements each of these steps. 

\begin{algorithm}[ht]
\begin{algorithmic}[1]
\Require $X$ (dataset)

\ \ $K'$ (number of subclusters, $K' \gg k^*$)

\ \ $M$ (minimum subcluster size)

\ \ $\alpha$ (significance level)
\vspace{0.2em}
\hrule
\vspace{0.2em}
\State \textbf{Overclustering:} Compute an initial overclustering of $X$ into $K'$ homogeneous subclusters lying in convex subregions. Eliminate small subclusters with less than $M$ datapoints, and determine the final overclustering partition $\mathcal{C}^{+}$ into $K\leq K'$ subclusters.
\State \textbf{Unimodal pair testing:} Apply a statistical test to determine whether the union of two subclusters admits unimodality (with significance level $\alpha$). 
\State \textbf{Clustering by subcluster aggregation:} Compute the final clustering partition $\mathcal{C}$ by determining the connected components of the unimodality graph $G$ of the overclustering $\mathcal{C}^{+}$.
\State \Return the locally unimodal clustering partition $\mathcal{C}$.
\end{algorithmic}
\caption{The \Uniforce general framework for locally unimodal clustering}
\label{alg:general}
\end{algorithm}

\subsection{Overclustering}\label{sec:overclustering}

The overclustering is an essential initial step for our bottom-up strategy, since we intent examine unimodality in local data regions and then to infer the larger scale cluster structure. More specifically, the overclustering step oversegments the unknown optimal partition $\mathcal{C}^{*}$, which we seek to discover, in $K' \gg k^{*}$ homogeneous subclusters lying in convex subregions (Fig.\,\ref{fig:K-plot-overclustering}).
%
Since $k^*$ is unknown, the hyperparameter $K$ of the method should be set to a sufficiently large value. The overclustering partition $\mathcal{C}^{+}$ will allow us to infer through a bottom-up aggregation the locally unimodal clusters (Fig.\,\ref{fig:K-plot-spunimodal}).

An algorithm of the $k$-means family can be employed to obtain a suitable overclustering partition containing homogeneous clusters lying in convex subregions. It should be noted that for a large number of clusters, the performance of the standard $k$-means algorithm deteriorates, as it is merely improbable to draw a good random initialization for many centers simultaneously. To mitigate this problem, we use instead the global $k$-means\texttt{++} algorithm~\cite{vardakas2022global}, which is an incremental variant that exhibits robust clustering performance for large numbers of clusters. More specifically, following the paradigm of global $k$-means~\cite{likas2003global}, global $k$-means\texttt{++} solves all intermediate clustering subproblems, for every $k \in [K]$, by adding one cluster center each time and selecting candidate center positions using the $k$-means\texttt{++}~\cite{arthur2006k} seeding procedure. 

Finally, in an additional preprocessing step (implemented by the function preprocess() in Alg.\,\ref{alg:method}), subclusters containing very few datapoints get eliminated, and their datapoints get redistributed to the remaining subclusters according to the $k$-means cluster assignment rule. As the number of subclusters increases, they become naturally more homogeneous, but also their cardinality becomes smaller. Since statistical testing lies at the heart of our methodology, namely the dip-test of unimodality~\cite{Hartigan1985}, we need to enforce a minimum allowed subcluster size, let that be $M$, to ensure the validity of the test. Empirical evidence suggests that a threshold to keep the dip-test more reliable is the sample size to be greater that $50$ datapoints~\cite{Ameijeiras-Alonso2019}. As we describe in detail next, our method tests pairs or subclusters whether they are jointly unimodal, which justifies setting the minimum subcluster size to $M=25$ datapoints.



\subsection{Unimodal pair testing}
\begin{figure}[t]
\centering
\begin{subfigure}{0.48\textwidth}
\centering
\includegraphics[scale=0.21]{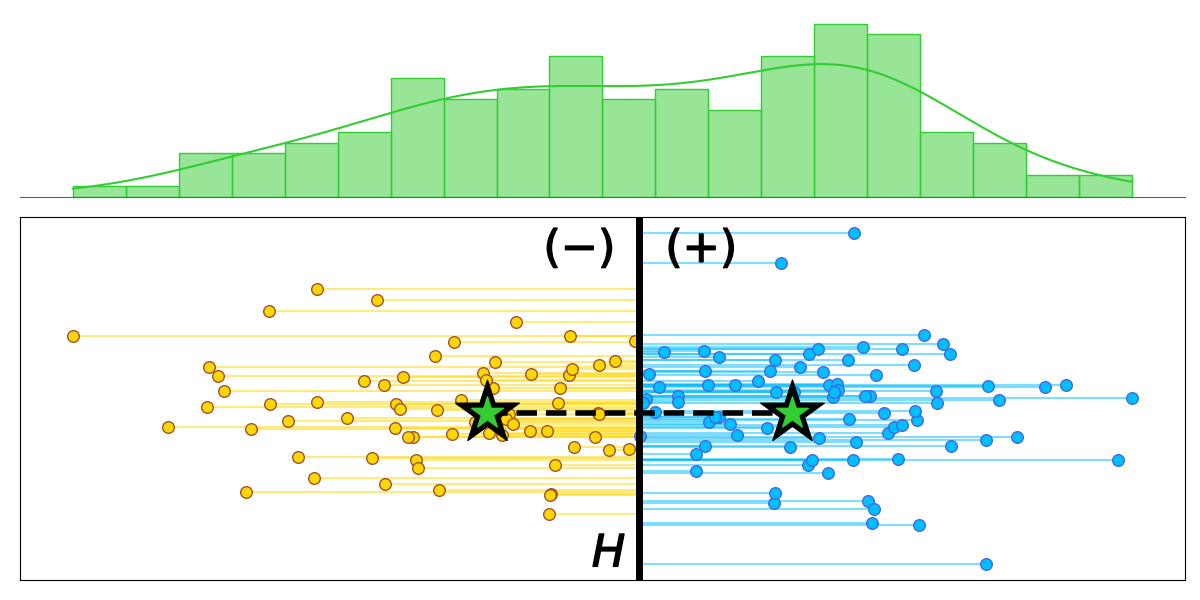}
\label{fig:unimodtest-unimodal}
\subcaption{\textbf{Unimodal case:} Two clusters forming a unimodal pair.}
\end{subfigure}
\begin{subfigure}{0.48\textwidth}
\centering
\includegraphics[scale=0.21]{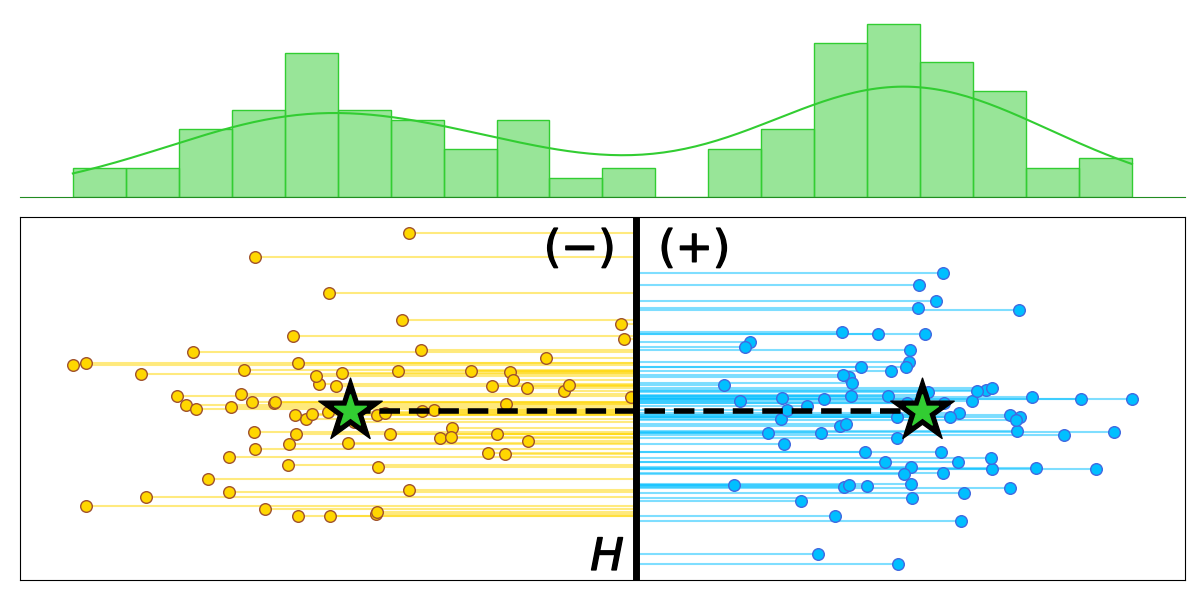}\subcaption{\textbf{Multimodal case:} Two clusters forming a multimodal pair.}
\label{fig:unimodtest-bimodal}
\end{subfigure}
\caption{
\textbf{Unimodal pair testing.}~Two subclusters, $c_i$ and $c_j$, appear in yellow and blue, respectively, and their centers are shown as stars. The dotted line connects the two centers, while the rigged line is its perpendicular bisecting hyperplane $H_{ij}$. Histograms present the density of the univariate set $P_{ij}$, with the point-to-hyperplane signed distances, which we test for unimodality using the dip-test. \textbf{a)}~Unimodal case: No density gap is observed between the subclusters and $P_{ij}$ is decided as unimodal. 
\textbf{b)}~Multimodal case: A considerable density gap is observed between the subclusters and $P_{ij}$ is decided as multimodal.} 
\label{fig:unimodtest}
\end{figure}


A step of major importance in our algorithm is the unimodal pair testing procedure that decides whether the union of two subclusters $c_{ij} = c_{i} \cup c_{j}$, where $c_i$, $c_j \in \mathcal{C}^{+}$, results in a unimodal cluster. We exploit the fact that we know the centers $\mu_i$ and $\mu_j$ of the subclusters tested for merging. We first define the vector $r_{ij} = \mu_i - \mu_j$ connecting these centers, and the perpendicular bisecting hyperplane $H_{ij}$ to $r_{ij}$ passing through its midpoint. Then, we consider the univariate set $P_{ij}$ that contains the signed distance of each point of $c_{ij}$ from the hyperplane $H_{ij}$. Finally, we apply the Hartigans' dip-test~\cite{Hartigan1985} for unimodality to $P_{ij}$ to decide whether it is unimodal with regards to a statistical significance level $\alpha$. 
Illustrations of a successful and an unsuccessful unimodal pair test (when it decides for unimodality, and multimodality, respectively) are provided in Fig.\,\ref{fig:unimodtest}. It is clear that if there is a density gap between the two subclusters, then the set $P_{ij}$ of signed distances 
will be multimodal, and therefore the dip-test is expected to reject unimodality. 

A sensitive aspect is that the dip-test may fail to reject unimodality when imbalanced modes are present. This means that if the number of datapoints belonging to one of the modes is much higher than the other, then it is quite probable for the test to decide for unimodality even when the two modes are quite well-separated. To address this issue, our approach uses a balanced subsample from the subclusters $c_i$ and $c_j$. More specifically, we use all the observations of the smaller subcluster and an equal uniform random subsample from the larger subcluster to produce a balanced set $c_{ij}$. Then we use the dip-test to for unimodality over the set $c_{ij}$. In order to get a more robust estimate, the we repeat this procedure (Monte Carlo experiment) for an odd number of times $L$, and decide the success or failure of the unimodal pair test based on the majority of the results of the $L$ experiments. The detailed unimodal pair test algorithm is presented in Alg.\,\ref{alg:2cluster-test}. 

\begin{algorithm}[t]
\begin{algorithmic}[1]
\Require $c_{i},c_{j}$ (two subclusters)

\ \ $L$ $\leftarrow 11$ (odd number of Monte Carlo simulations)

\ \ $\alpha \leftarrow 0.001$ (significance level)

\vspace{0.2em}
\hrule
\vspace{0.2em}
\State $\mu_i \leftarrow \frac{1}{|c_i|}\sum\limits_{x \in c_i} x$; \ \ $\mu_j \leftarrow \frac{1}{|c_j|}\sum\limits_{x \in c_j} x$ \Comment{{\footnotesize The centers of the two subclusters}}
\State Find the perpendicular bisecting hyperplane $H_{ij}$ to the vector $r_{ij} = \mu_j - \mu_i$ connecting the two centers
\State Let $s \leftarrow \min\{|c_i|,|c_j|\}$ \Comment{{\footnotesize Minimum cardinality of the two sets}}
\State Initialize $v$ as a zero vector with $L$ elements \Comment{{\footnotesize Votes for unimodality}}
\ForAll{$l \in [L]$}
\State Create $c'_{i}$ by sampling $s$ elements uniformly at random without replacement from $c_{i}$
\State Create $c'_{j}$ by sampling $s$ elements uniformly at random without replacement from $c_{j}$         
\State Create the union set $c'_{ij} \leftarrow c'_{i} \cup c'_{j}$
\State Compute the set $P_{ij}$ with the signed distances from $H_{ij}$ for all samples in $c'_{ij}$
\State $p \leftarrow \text{dip-test}(P_{ij})$ \Comment{{\footnotesize The $p$-value of the Hartigans' dip-test for unimodality~\cite{Hartigan1985}}}
\State $v[l] \leftarrow  \mathds{1}\{p \geq \alpha\}$ \Comment{{\footnotesize Store the vote against or for unimodality, $0$ or $1$ respectively}}
\EndFor
\State $m \leftarrow \mathds{1}\big\{\textstyle\sum_{l=1}^L v[l] > \frac{L}{2}\big\}$ \Comment{{\footnotesize Compute the majority vote, either $0$ or $1$}}
\State \Return $m$  
\end{algorithmic}
\caption{Unimodality test for a pair of subclusters 
\Comment{\footnotesize Called as \myvar{unimodalPair()} in Alg.\,\ref{alg:method})}}
\label{alg:2cluster-test}
\end{algorithm}




\subsection{Finding connected components}
The application of the unimodal pair testing procedure on subcluster pairs provides the unimodality graph of the initial overclustering partition. Finding the connected components of the unimodality graph is the next step in our methodology. We choose to represent each connected component by a spanning tree, which is the minimum structure required. Each spanning tree of the unimodality graph, called \emph{unimodal spanning tree}, represents a maximal locally unimodal cluster, and the \emph{unimodality spanning forest} provides the overall clustering partition. The \Uniforce algorithm computes the unimodality spanning forest and uses an online graph construction procedure to minimize the number of the required unimodality tests. Specifically, we make two relaxations that make the computation much more efficient without affecting the clustering result. First, we remark that any spanning tree (not necessarily the minimum one) connecting the same set of subclusters of a given overclustering would produce the same clustering partition. Second, since unimodality extends locally across neighboring subclusters, we can use the proximity of between pairs of subclusters as a preference for the order in which pairs shall get tested for unimodality. 

The exhaustive computation of the unimodality graph for $K$ subclusters would require $\frac{K(K-1)}{2}$ unimodality tests. In order to reduce the computational cost, we use a simpler strategy by exploiting the above-mentioned preference order for testing pairs of (closely) neighboring subclusters. We start with a complete distance graph $G$, whose vertices are the subclusters and the edge weights $W$ are the $\frac{K(K-1)}{2}$ pairwise Euclidean distances (using other alternatives adapted to the data is possible) between the centers of the subclusters. Then, we consider the proximity of two subclusters as an indicator for the possibility that unimodality extends 
across those subclusters, and hence we test pairs of subclusters in pairwise proximity order. 

The unimodal spanning forest approximation $F$ is computed over $G$ by also testing for spanning unimodality between pairs of vertices. Algorithmically, we compute $F$ using a modification of the classical Kruskal's algorithm~\cite{Kleinberg+Tardos:06a}. Initially, $F$ consists of $K$ trees, each with only one vertex. The edges of $G$, sorted in ascending weight order, indicate the order in which unimodality between pairs of vertices should be tested. When a test for a pair of vertices is successful, we add an edge in $F$ connecting those vertices. Each time this occurs, the number of trees (also clusters of the partition) is reduced by $1$. While traversing this list of edge weights, we skip pairs of vertices that are already in the same spanning tree of $F$. Operating in the described way minimizes the number of unimodality tests that need to be performed without affecting the final clustering result.

\begin{algorithm}[t]
\begin{algorithmic}[1]
\Require $X$ (dataset)
 
\ \ $K' \leftarrow 50$ (number of subclusters)

\ \ $M \leftarrow 25$ (minimum subcluster size)

\ \ $L \leftarrow 11$ (odd number of Monte Carlo Simulations)

\ \ $\alpha \leftarrow 0.001$ (significance level)
\vspace{0.2em}
\hrule
\vspace{0.2em}

\State $\{\mathcal{C}^{+}=\{c_i\}_{i\in[K]}, \mu = \{\mu_i\}_{i\in[K]}\} \leftarrow$ preprocess$($global $k$-means\texttt{++}$(X, K'), M)$ \Comment{{\footnotesize Overclustering into $K$}subclusters,}
\Statex \Comment{{\footnotesize with more than $M$ datapoints each}}
   
\State Consider the subclusters' centers $\{\mu_i\}$ as graph vertices $\{V_{i}\}$, $i \in [K]$ 
\State Compute the distance graph $G$ with edge weights $W_{ij} = \text{dist}(\mu_{i}, \mu_{j})$, $i,j \in [K]$ \Comment{{\footnotesize $\text{dist}(\cdot, \cdot)\leftarrow$ Euclidean distance}}
\State Initialize the unimodality spanning forest $F$ with $K$ singleton trees, one for each $V_i \in G$, $i \in [K]$
\For{\textbf{each} edge $(V_i, V_j) \in G$ in ascending order of distance $W_{ij}$} \Comment{{\footnotesize Kruskal's algorithm}}
\If {\myvar{inDifferentTrees}$(F, V_{i}, V_{j})$ \textbf{and }\myvar{unimodalPair}$(c_{i}, c_{j}, L, \alpha)$}
\State Add the edge $(V_i, V_j)$ in $F$
\EndIf
\EndFor
\For{\textbf{each} unimodal spanning tree $T_{j} \in F$}  \Comment{{\footnotesize Consider a cluster from each tree}}
\State Create the cluster $C_j$ with the vertices $V_i \in T_{j}$ \Comment{{\footnotesize Gather all datapoints of those subclusters}}
\EndFor
\State \Return the locally unimodal cluster partition $\mathcal{C} = \{C_{1}, \ldots, C_{|F|}\}$, and the estimated $k = |F|$
\end{algorithmic}
\caption{The \Uniforce algorithm for clustering and estimation of the number of clusters}
\label{alg:method}
\end{algorithm}

\subsection{Complexity analysis}\label{sec:complexity}

The methods that rely on the dip-test, including \Uniforce, can benefit from a dramatic acceleration of the statistical tests by making use of a lookup table with precomputed results for the bootstrap dip statistic of Uniform samples\footnote{In our implementation we used the repository that is available at: \href{https://pypi.org/project/diptest/}{https://pypi.org/project/diptest/}.}, instead of analytically computing it. The lookup table stores precomputed dip statistics over a grid of sample sizes and significance levels. The grid needs to be sufficiently dense for the scale of the treated problem. Other interesting approaches to accelerate the dip-test have started to appear in the literature \cite{accelareteDipTest2023}, showing additional possible directions for future refinements.

For the overclustering step, a computationally cheap choice is to use $k$-means\texttt{++}~\cite{arthur2006k}. However, in Sec.\,\ref{sec:overclustering}, we justified the use of the global $k$-means\texttt{++} algorithm~\cite{vardakas2022global} by the fact that it takes $\bigO(QKNd)$ time, where $Q$ is the number of candidates examined per incremental iteration to initialize the new cluster, $K$ is the desired number of subclusters, and $N$ is the size of the dataset and $d$ is the dimensionality of the data. The number of candidates $Q$ should be $\bigO(1)$, e.g.~between $10$ and $20$. Thus, we can say that the overclustering step takes $\bigO(KNd)$ time. Regarding preprocessing, removing the very small subclusters from the overclustering partition and reassigning their elements takes $\bigO(KN)$ time. 

In the subsequent analysis, we delve into the overall time complexity associated with the invocations of the \myvar{unimodalPair} function. The time complexity to compute the dip statistic of a dataset of size $n$ using Hartigans' dip-test, is $\bigO(n)$~\cite{Hartigan1985} provided that the values are sorted. However, since sorting is necessary, the time complexity for calling once the \myvar{unimodalPair} function is $\bigO(n\log{n})$. We can show that the total time complexity of computing the dip statistic of all unimodal pairs is of $\bigO(N\log{N})$:
\begin{equation*}
\label{eq:complexity}
\sum_{\begin{subarray}{c}i,j\in [K],\,i<j\end{subarray}}{(|V_{i}|+|V_{j}|)\log(|V_{i}|+|V_{j}|)}\leq\Bigg(
\sum_{i,j\in [K]}{(|V_{i}|+|V_{j}|)}\Bigg)
\log{N}
\leq 2N\log{N}.
\end{equation*}
Note that the number of tests $M$ is supposed to be $\bigO(1)$, e.g.~between $1$ and $11$. The computation of the spanning forest $F$ takes $\bigO(K^{2}\log{K})$ time. Constructing the final clustering takes $\bigO(N)$ time.  Thus, the \emph{total time complexity} of \Uniforce algorithm is $\bigO(KNd + KN + N\log{N} + K^2\log{K})$.

\subsection{Discussion}\label{sec:discussion}
Based on the previous presentation, the \Uniforce algorithm can be seen as a \emph{single-link agglomerative} approach over an initial overclustering of the data into subclusters. The data aggregation in such approaches acts over those subclusters: at any point throughout the process, a link between two subclusters decides for the merging of the clusters (possibly containing several other subclusters) in which they belong. In that sense, the aggregation is controlled by a local merging criterion between subclusters and does not impose any condition over the properties of the whole set of the two merged clusters. This latter feature is in contrast to other agglomerative clustering approaches. 

It is important to stress that the identification of arbitrary-shaped clusters can be enabled mainly through data aggregation using local cluster-merging criteria. This type of criteria allow a cluster to be defined over a region that extends for as long as a chosen statistical property holds, without committing to any global cluster shape. This general strategy has been used extensively in the literature, either formalized as the aforementioned single-link agglomerative approach, or as a density-based approach (DBSCAN and variants). The latter grows a cluster region toward a neighboring area by using a local criterion accounting for the local data density at that area. An advantage of this framework is that the cluster-growing criterion decides also the termination of the process, hence also estimates the number of clusters. This is not the case for the typical agglomerative clustering, where an external criterion is usually needed to terminate the process (e.g.~the Silhouette Coefficient), which is not associated to the employed cluster-merging criterion.

Despite the numerous clustering methodologies proposed in this vein, the main limitation of most existing works is that they use heuristic cluster-merging/-growing criteria whose parameters are particularly difficult to tune (e.g.~relying on the distances between clusters, simple features of local geometry, cluster variance, local density level, etc.). The novelty of \Uniforce lies in the use of a new local property controlling the merging of the clusters, which is the \emph{local unimodality} tested between pairs of neighboring subclusters. This is a statistically sound way for cluster-merging that conditions the local density shape and identifies what we defined as \emph{locally unimodal clusters}. Moreover, our cluster-merging criterion determines automatically the number of clusters of the final partition, as it terminates the agglomeration without involving external stopping criteria. The technical means allowing to test efficiently for local unimodality is our unimodal paired test, which is tailored to the context of our clustering algorithm and is among the main contributions of this work.

Regarding deploying \Uniforce in practice, setting a suitable number of initial subclusters $K$ and finding a quality overclustering needs attention (Sec.\,\ref{sec:overclustering}). However, this requires way less sophistication compared to solving the final task, as multiple $K$ values may lead to similar final clustering partitions (for instance, see Fig.\,\ref{fig:sensitivity-plots}). In our experiments a fixed value $K=50$ was used in all datasets. Moreover, since our methodology relies on statistical testing, the subclusters need to have sufficient data density. This may hinder the identification of very small clusters (e.g.~with less than $50$ datapoints). 

\section{Experimental evaluation}\label{sec:experiments}

\subsection{Experimental setup}

\renewcommand{\arraystretch}{1}
\begin{table}[pt!]\footnotesize
\centering
\caption{The real datasets used in our experiments. $N$ is the number of data instances, $d$ is the dimensionality, and $k$ denotes labeled classes that we consider as the ground-truth number $k^*$. With `\texttt{*}', we mark an embeddings dataset obtained by training an autoencoder on the original dataset.}
\label{tab:data}
\resizebox{0.9\columnwidth}{!}{
\begin{tabular}{lllrrrc}
\toprule
\textbf{Dataset}&\textbf{Type}&\textbf{Description}&{${N}$}&\textbf{${d}$}&\textbf{${k}$}&\textbf{Source}\\
\midrule
Waveform-v1&Vector&Waveforms with multiple attributes&5000&21&3&\cite{Dua:2019}\\
Mice Protein Expression&Tabular&Expression levels of proteins&1080&77&8&\cite{Dua:2019}\\
Optdigits&Image&Handwritten digits&5620&$8\times8$&10&\cite{Dua:2019}\\
Pendigits&Time-series&Handwritten digits&10992&16&10&\cite{Dua:2019}\\
EMNIST Balanced Digits \ \texttt{*}&Vector&Handwritten digits&28000&10&10&\cite{cohen2017emnist}\\
EMNIST MNIST\quad\ \,\,\ \qquad  \texttt{*}&Vector&Handwritten digits&70000&10&10&\cite{cohen2017emnist}\\
EMNIST Balanced Letters\,\texttt{*} &Vector&Handwritten letters (A-J) &28000&10&10&\cite{cohen2017emnist}\\
Isolet&Spectral&Speech recordings pronouncing letters&7797&617&26&\cite{Dua:2019}\\
TCGA&Tabular&Cancer gene expression profiles&801&20531&5&\cite{Dua:2019}\\
Complex $2$D (synthetic)&Vector& Multiple structures inside a ring (Fig.\,\ref{fig:K-plot})&5000&2&6&ours\\
\bottomrule
\end{tabular}}
\end{table}

\textbf{Datasets.}~Tab.\,\ref{tab:data} summarizes the benchmark datasets that we used for experimental evaluation, which vary in size $N$, dimensions $d$, number of clusters $k$ (this is the number of labeled classes that we consider as the ground-truth value $k^*$), data type, and domain of origin. 

The datasets Optigits, Pendigits, EMNIST MNIST (EMNIST-M), and EMNIST Balanced Digits (EMNIST-BD) comprise handwritten digits, with $10$ classes corresponding to the digits from $0$ to $9$. Optigits consist of images with a resolution of $8\times8$, while EMNIST-M and EMNIST-BD contain images with a higher resolution of $28\times28$. In contrast, Pendigits' data instances are represented by $16$-dimensional vectors containing pixel coordinates. EMNIST-BL is a dataset with handwritten letters, with capital and non-capital characters, from which we selected the $10$ classes corresponding to the letters A to J, that account for $28000$ datapoints. The Isolet dataset is a collection of speech recordings containing the sound samples of spoken letters, represented by vectors of $617$ spectral coefficients extracted from the speech signal. The TCGA is a collection of gene expression profiles obtained from RNA sequencing of various cancer samples. It includes $801$ data instances, clinical information, normalized counts, gene annotations, and $6$ cancer types' pathways. The Mice Protein Expression dataset consists of the expression levels of 77 proteins/protein modifications that produced detectable signals in the nuclear fraction of the cortex. It includes $1080$ datapoints and $8$ eight classes of mice based on genotype, behavior, and treatment features. The Waveform-v1 consists of 3 classes of generated waves with $5000$ datapoints. Each class is generated from a combination of 2 of 3 `base' waves. 

EMNIST is an extended and more challenging MNIST dataset. Due to the high complexity of the three EMNIST versions, we used these datasets after creating a high-quality data embedding via an Autoencoder (AE). The architecture of the convolutional AE is highly used in literature for clustering purposes~\cite{guo2018deep,ren2020deep,guo2021deep}. Specifically, the encoder part consists of $3$ convolutional layers with channel numbers $32$, $64$, and $128$, and kernel sizes of $5\times5$, $5\times5$, and $3\times3$, respectively. This is followed by a two-layer MLP with $384$ and $10$ neurons, respectively. The decoder part of the AE is symmetrical with the encoder. LeakyReLU~\cite{maas2013rectifier} activates all intermediate layers of the AE with a slope equal to $0.1$. We trained the AE for $100$ epochs using the Adam optimizer~\cite{kingma2014adam} with a constant learning rate of $0.001$, batch size of $256$ and with the default setting of $\beta_1 = 0.9$ and $\beta_2 = 0.999$. For the Mice Protein Expression dataset, we applied one-hot encoding to manage categorical values and to address the few missing data; we imputed the missing values by utilizing the mean values for each column. As a preprocessing step, we used min-max normalization to map the attributes of each dataset to the $[0, 1]$ interval to prevent attributes with large value ranges from dominating the distance calculations, and to also avoid numerical instabilities in the computations~\cite{celebi2013comparative}. 

The $2$-dimensional Complex $2$D is the only synthetic dataset that we use in the first experimental part. It contains multiple clusters inside a ring (see  Fig.\,\ref{fig:K-plot}), some of which are non-convex and pairwise non-linearly separable. More additional experiments on synthetic datasets are presented in Sec.\,\ref{sec:synthetic_data}.

\textbf{Compared clustering methods.}~The performance of the \Uniforce algorithm is compared with several methods that perform clustering and automatic estimation of the number of clusters. The most related category of methods are those using statistical tests: X-means~\cite{x-means-2000}, G-means~\cite{g-means-2003}, PG-means~\cite{pg-means-2006}, dip-means~\cite{dip-means-2012}, and projected dip-means~\cite{pdip-means-2018} (pdip-means). Additionally, we considered approaches that do not rely on statistical tests in their optimization procedure, such as HDBSCAN~\cite{campello2015hierarchical}, RCC~\cite{shah2017robust}, and Mean Shift~\cite{comaniciu2002mean}. HDBSCAN is a method that performs DBSCAN~\cite{ester1996density} over varying epsilon values and integrates the results to find a clustering partition that gives the best stability over epsilon. Since, by design, we stay in the original data space, we do not consider approaches that integrate embedding and clustering, such as deep clustering methods. In all experiments, we fixed our hyperparameters to $K=50$ (number of initial subclusters), $\alpha=0.001$ (statistical significance level), $M=25$ (minimum subcluster size), and $L=11$ (number of Monte Carlo executions). 

\textbf{Evaluation measures.}~For evaluating how well a clustering partition matches the ground-truth label information, we compute the \emph{Adjusted Mutual Information} (AMI)~\cite{JMLR:v11:vinh10a} measure that takes values from $0$ to $1$, and is defined as:
\begin{equation*}
\text{AMI}(Y, C) = \frac{I(Y,C) - \mathbb{E}[I(Y,C)]}{\max\{H(Y), H(C)\} - \mathbb{E}[I(Y,C)]},
\end{equation*} 
where $Y$ denotes the vector of ground-truth labels, $C$ denotes the vector of cluster labels produced by a clustering algorithm, $I$ is the Mutual Information measure, $H$ the entropy of a partition (either the ground-truth or the produced one), and $\mathbb{E}[\cdot]$ is the expected value. Higher AMI values indicate that a clustering partition matches better with the ground-truth labels. We report the average values for $k$ and AMI obtained from $30$ executions of each method on each dataset. Care is needed when interpreting results concerning the estimation of the number of clusters since a correct estimation of $k$ does not necessarily imply a correct clustering solution. Therefore, the AMI measure and $k$ should simultaneously be considered to draw safer conclusions.

\subsection{Numerical results}
\begin{table}[t]\footnotesize
    \centering
    \caption{Summary of the experimental results. The best values per dataset are shown in bold. Cases marked by $\dagger$ and $\ddagger$ indicate experiments that failed due to memory/time and method constraints, respectively.}
    \label{tab:res}
    \resizebox{0.85\columnwidth}{!}{
    \begin{tabular}{>{}rr@{$\pm$}rrrr@{$\pm$}rrrr@{$\pm$}rrrr@{$\pm$}rrrr@{$\pm$}rr}
        \toprule 
        &\multicolumn{3}{c}{\textBF{Optdigits}}&\ &\multicolumn{3}{c}{\textBF{Pendigits}}&\ &\multicolumn{3}{c}{\textBF{EMNIST-BD}} &\ &\multicolumn{3}{c}{\textBF{EMNIST-M}}& \ &\multicolumn{3}{c}{\textBF{EMNIST-BL}}\\
        
        \cmidrule{2-20}
        &\multicolumn{2}{c}{$k$}&\multicolumn{1}{c}{AMI}&&\multicolumn{2}{c}{$k$}&\multicolumn{1}{c}{AMI}&&\multicolumn{2}{c}{$k$}&\multicolumn{1}{c}{AMI}&&\multicolumn{2}{c}{$k$}&\multicolumn{1}{c}{AMI}&&\multicolumn{2}{c}{$k$}&\multicolumn{1}{c}{AMI}\\
        \cmidrule{2-4}\cmidrule{6-8}\cmidrule{10-12}\cmidrule{14-16}\cmidrule{18-20}

        X-means&422&9&0.34&&1472&19&0.25&&357&10&0.37&&576&15&0.35&&286&9&0.37\\
        G-means&57&5&0.53&&184&11&0.44&&120&5&0.44&&265&8&0.39&&139&11&0.41\\
        PG-means&\multicolumn{2}{r}{$\ddagger$}&$\ddagger$&&25&3&0.58&&42&4&0.58&&48&5&0.58&&44&4&0.56\\
        dip-means&4&0&0.38&&\textbf{9}&1&0.62&&\multicolumn{2}{r}{7 or 8}&0.60&&9&0&0.75&&5&0&0.60\\
        pdip-means&12&1&0.69&&12&1&0.69&&7&0&0.55&&5&0&0.61&&3&0&0.45\\
        Mean Shift&73&0& 0.63&&17&0&0.65&&\multicolumn{2}{r}{$\dagger$}&$\dagger$&&\multicolumn{2}{r}{$\dagger$}&$\dagger$&&\multicolumn{2}{r}{$\dagger$}&$\dagger$\\
        HDBSCAN&9&0&0.48&&28&0&0.69&&12&0&0.40&&\textbf{10}&0&0.45&&3&0&0.02\\
        RCC&19&0&\textbf{0.87}&&46&0& 0.75&&82&0&0.74&&\multicolumn{2}{r}{$\dagger$}&$\dagger$&&82&0&0.52\\
        \Uniforce&\textbf{11}&1&0.85&&17&1&\textbf{0.78}&&\textbf{10}&1&\textbf{0.86}&&13&1&\textbf{0.84}&& \textbf{12} & 1 & \textbf{0.74}\\
		\midrule
		Ground-truth&\multicolumn{2}{c}{10}&-&&\multicolumn{2}{c}{10}&-&&\multicolumn{2}{c}{10}&-&&\multicolumn{2}{c}{10}&-&&\multicolumn{2}{c}{10}&-\\
		\midrule
        \addlinespace    
        &\multicolumn{3}{c}{\textBF{Waveform-v1}}&\ &\multicolumn{3}{c}{\textBF{Mice Protein}}&\ &\multicolumn{3}{c}{\textBF{Isolet}}&\ &\multicolumn{3}{c}{\textBF{TCGA}}&\ &\multicolumn{3}{c}{\textBF{Complex 2D}}\\
        \cmidrule{2-20}
        &\multicolumn{2}{c}{$k$}&\multicolumn{1}{c}{AMI}&&\multicolumn{2}{c}{$k$}&\multicolumn{1}{c}{AMI}&&\multicolumn{2}{c}{$k$}&\multicolumn{1}{c}{AMI}&&\multicolumn{2}{c}{$k$}&\multicolumn{1}{c}{AMI}&&\multicolumn{2}{c}{$k$}&\multicolumn{1}{c}{AMI}\\
        \cmidrule{2-4}\cmidrule{6-8}\cmidrule{10-12}\cmidrule{14-16}\cmidrule{18-20}
    
        X-means&10&0&0.32&&244&2&0.27&&233&4&0.49&&20&1&0.51&&1&0&0.00\\
        G-means&12&0&0.31&&32&1&0.71&&101&6&0.57&&\multicolumn{2}{r}{$\ddagger$}&$\ddagger$&&96&1&0.28\\
        PG-means&4&1&0.45&&\multicolumn{2}{r}{$\ddagger$}&$\ddagger$&&\multicolumn{2}{r}{$\ddagger$}&$\ddagger$&&\multicolumn{2}{r}{$\dagger$}&$\dagger$&&23&2&0.42\\
        dip-means&4&0&0.42&&5&0&0.67&&4&0&0.30&&2&0&0.36&&26&1&0.32\\
        pdip-means&1&0&0.00&&\multicolumn{2}{r}{9 or 10}&\textbf{0.96}&&16&1&0.52&&\multicolumn{2}{r}{$\ddagger$}&$\ddagger$&&\textBF{6}&0&0.37\\
        Mean Shift&2&0&0.34&&6&0&0.73&&\multicolumn{2}{r}{$\dagger$}&$\dagger$&&\multicolumn{2}{r}{$\dagger$}&$\dagger$&&7&0&0.30\\
        HDBSCAN&4&0&0.85&&11&0&0.93&&3&0&0.02&&5&0&0.55&&7&0&0.90\\
        RCC&\textbf{3}&0&\textbf{1.00} &&54&0&0.52&&12&0&0.53&&8&0&0.84&&498&0&0.09\\
        \Uniforce&\textbf{3}&0&\textbf{1.00}&&\textbf{8}&0&0.93&&\textbf{27}&2&\textbf{0.71} &&\multicolumn{2}{r}{\textbf{5 or 6}}&\textbf{0.87}&&\textbf{6}&0&\textbf{0.98}\\
		\midrule
		Ground-truth&\multicolumn{2}{c}{3}&-&&\multicolumn{2}{c}{8}&-&&\multicolumn{2}{c}{26}&-&&\multicolumn{2}{c}{5}&-&&\multicolumn{2}{c}{6}&-\\
        \bottomrule
    \end{tabular}}
\end{table}

\begin{figure}[t]
\centering
\begin{subfigure}{0.8\textwidth}
\centering
\includegraphics[width=\linewidth]{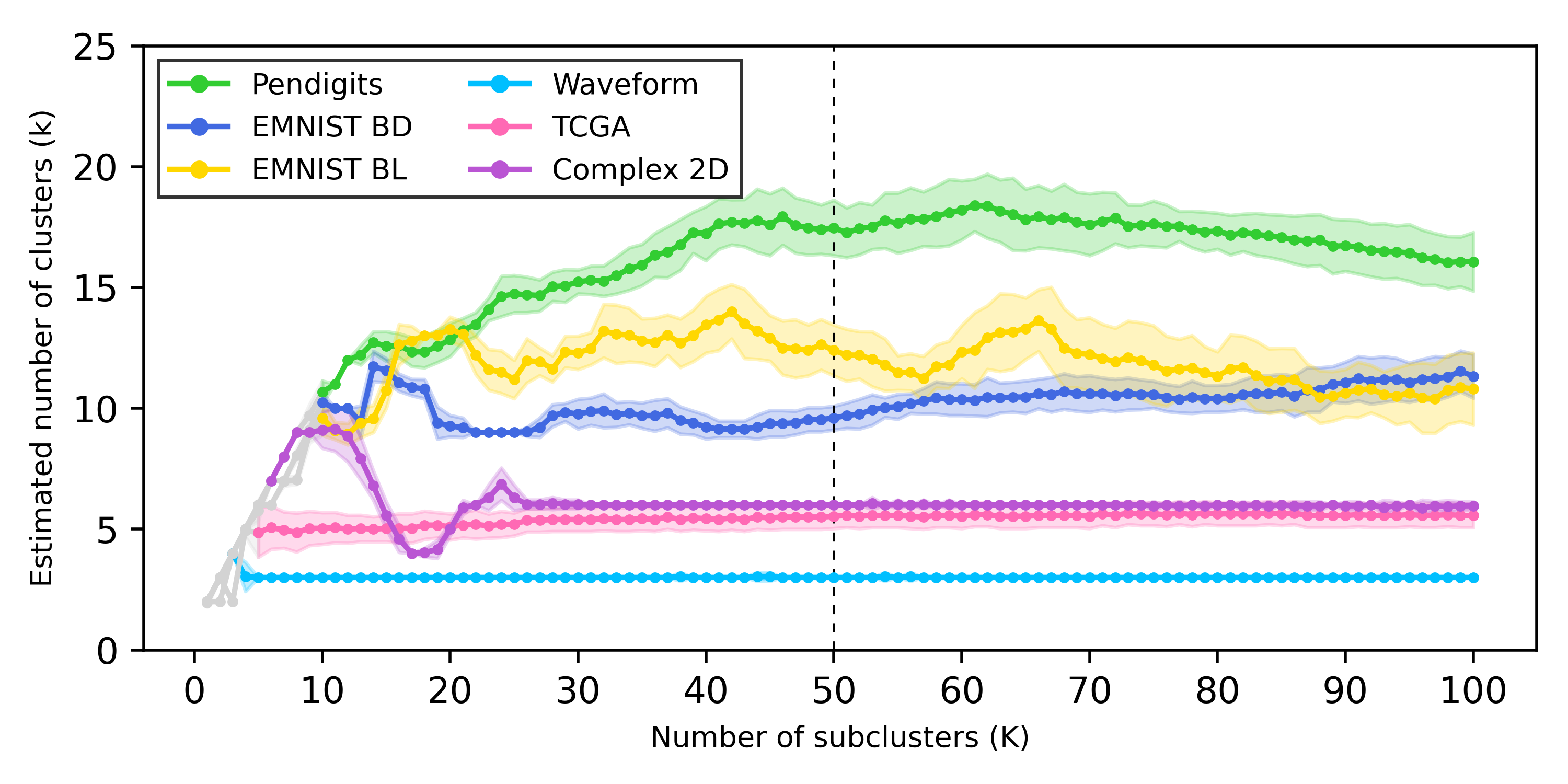}
\caption{$k$ vs $K$: Estimation of the number of clusters ($k$) by the \Uniforce method as a function of the number of starting subclusters ($K$).}
\label{fig:K_vs_k_sensitivity-plots}
\end{subfigure}
\vspace{1mm} 
\begin{subfigure}{0.8\textwidth}
\centering
\includegraphics[width=\linewidth]{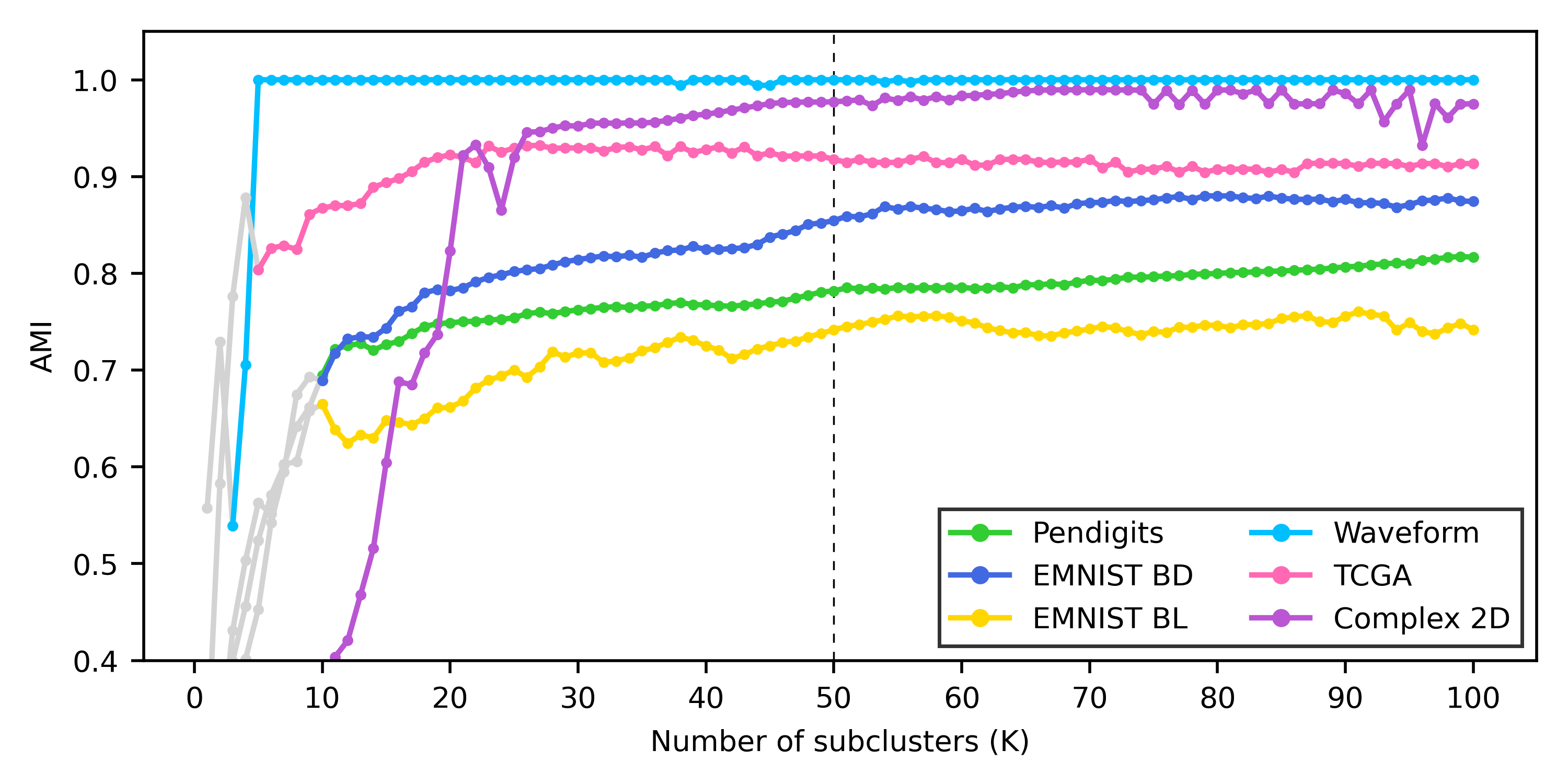}
\caption{AMI vs $K$: Clustering performance (AMI score) by the \Uniforce method as a function of the number of starting subclusters ($K$).}
\label{fig:AMI_vs_K_sensitivity-plots}
\end{subfigure}
\caption{Comparison of clustering results. The gray part of each curve corresponds to clustering solutions where $K < k^*$, for the associated dataset.}
\label{fig:sensitivity-plots}
\end{figure}
The experimental results are summarized in Tab.\,\ref{tab:res}.
First, we empirically confirm results known in the literature, that top-down methods such as X-means and G-means fail to capture the structure a dataset unless their assumptions are rather true. X-means exploits the BIC criterion, while G-means relies on statistical tests for Gaussianity. Their estimations of the number of clusters is one or even two orders of magnitude higher than the actual number of clusters in the data. 
Dip-means and pdip-means are also top-down approaches but they rely on unimodality tests and consistently outperform X-means and G-means  providing better estimations for the number of clusters. Therefore, it is evident that the unimodality-based methods perform better compared to methods that make `stricter' assumptions, such as Gaussianity. 
However, on datasets containing many clusters, both dip-means and pdip-means terminate too early and fail to reasonably estimate the number of clusters. There are two distinct sources for this shortcoming: first, their approach for testing dataset unimodality (such as the dip-dist, which is a 'meta-test', and variations of it) is not very effective in the multivariate setting and, since the methods operate in a top-down fashion, there is high chance for false positive identification of unimodality; second, the structure of the true clusters may be complex, thus the classical unimodality assumption may not be valid. This is what we aim to capture with the proposed locally unimodal cluster definition. In addition, Mean Shift performed poorly in all datasets. HDBSCAN gave promising results in Waveform-v1 and Mice Protein, while its performance was poor in the remaining datasets due to low AMI value or false $k$ detection. The RCC method produced satisfactory results on the Optdigits, Waveform-v1 and TCGA datasets, yet it failed on the rest of the benchmark datasets because the $k$ estimation is far from the ground truth labeling.

The \Uniforce algorithm performed satisfactorily on all benchmark datasets. It produced high-quality estimates of the number of clusters and high-quality clusterings with a high AMI on all datasets. More specifically, on the EMNIST-BD, EMNIST-BL, Waveform-v1, Isolet, TCGA, and Complex 2D datasets, \Uniforce outperformed the other methods in both estimating $k$ and providing clustering solutions with high AMI. In addition, the \Uniforce method had the best solutions for $k$ in the Optdigits and Mice Protein datasets, while it was highly competitive in AMI. Finally, in the Pendigits and EMNIST-M datasets, the method had a very high AMI score with a good performance on $k$, where it overestimated the ground truth labeling by a small margin. However, it should be noted that the detectable cluster structure $k$ is not always aligned with the number of classes labeled in the dataset. 


Although in all previous experiments we considered the same number of initial clusters $K=50$, in order to study the influence of the overclustering resolution in the performance of our method, we conducted additional experiments by varying the associated hyperparameter $K=1,\ldots,100$. The plots in Fig.\,\ref{fig:K_vs_k_sensitivity-plots}, showcase that the \Uniforce method is quite robust with respect to the parameter $K$, except for the EMNIST-BL dataset, where the final number of clusters $k$ increases as the initial number of clusters $K$ increases. This may be an indication of the existence of a large number of substructures (much higher than the number of ground truth classes $k^{*}$). Finally, Fig.\,\ref{fig:AMI_vs_K_sensitivity-plots} provides a detailed view over the effect that the value of $K$ has on the AMI measure. For completeness, we included for each curve a first part appearing in gray, which corresponds to when $K < k^*$, and therefore the cases where the initialization is not not an overclustering, but rather an underclustering.

\begin{figure}[ht]
  \centering
  \begin{subfigure}[b]{0.2\linewidth}
    \includegraphics[width=\linewidth]{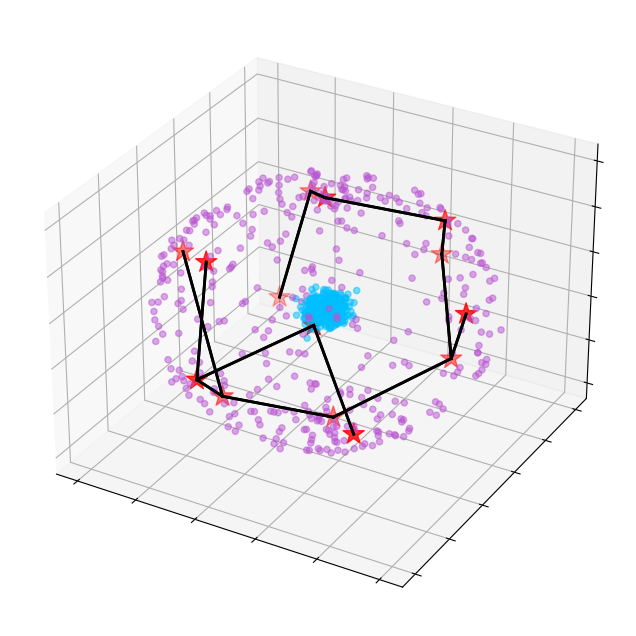}
  \end{subfigure}
  \hfill
  \begin{subfigure}[b]{0.2\linewidth}
    \includegraphics[width=\linewidth]{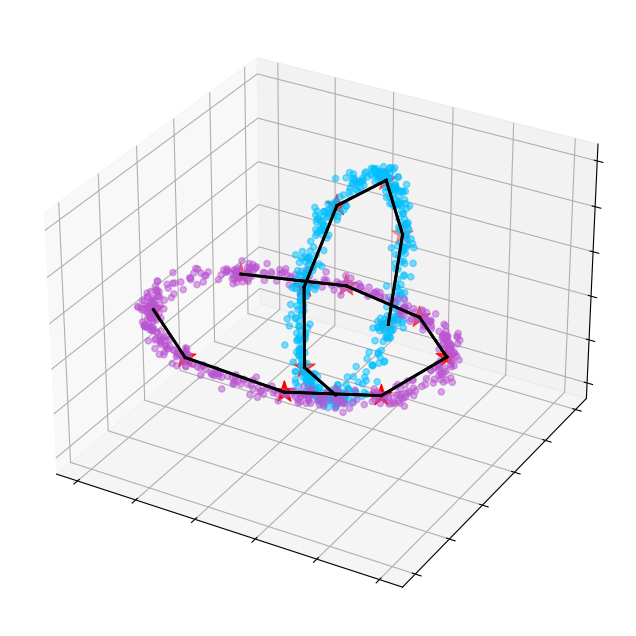}
  \end{subfigure}
  \hfill
  \begin{subfigure}[b]{0.2\linewidth}
    \includegraphics[width=\linewidth]{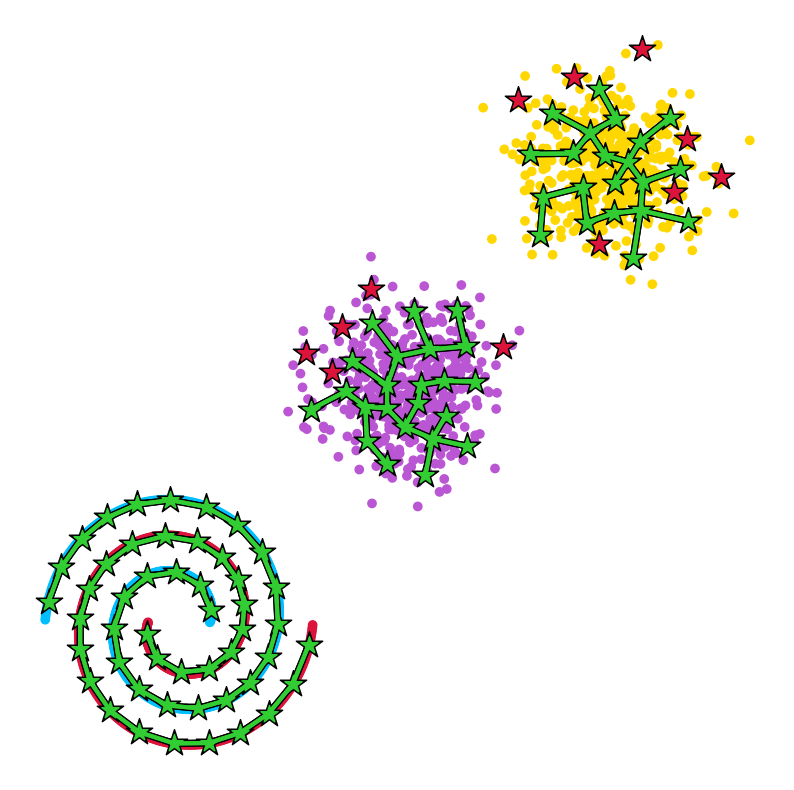}
  \end{subfigure}
  \hfill
	\begin{subfigure}[b]{0.2\linewidth}
    \includegraphics[width=\linewidth]{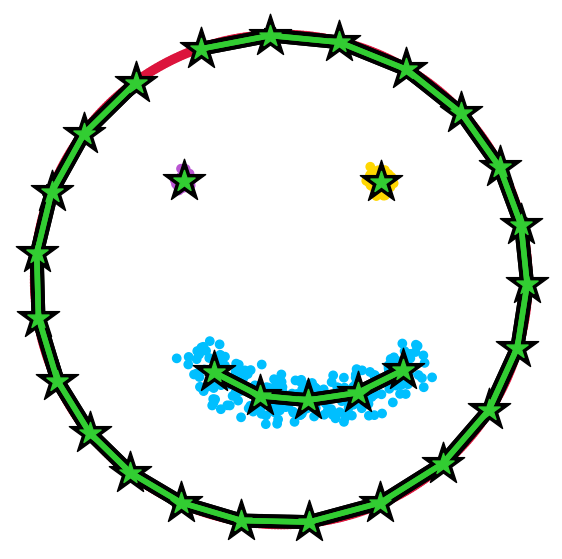}
  \end{subfigure}
  \\
  \begin{subfigure}[b]{0.2\linewidth}
    \includegraphics[width=\linewidth]{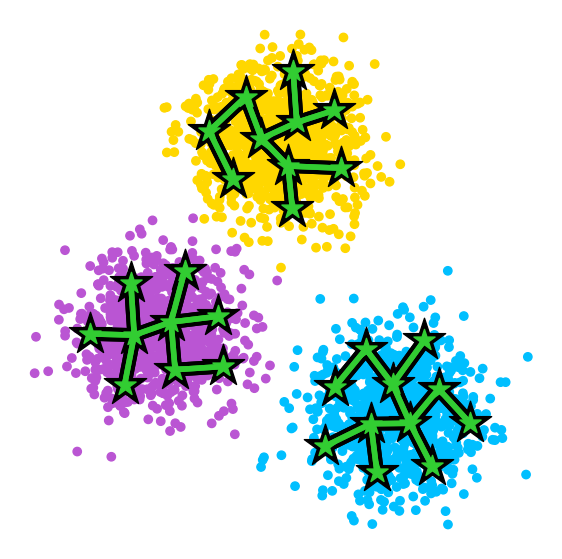}
  \end{subfigure}
  \hfill
  \begin{subfigure}[b]{0.2\linewidth}
    \includegraphics[width=\linewidth]{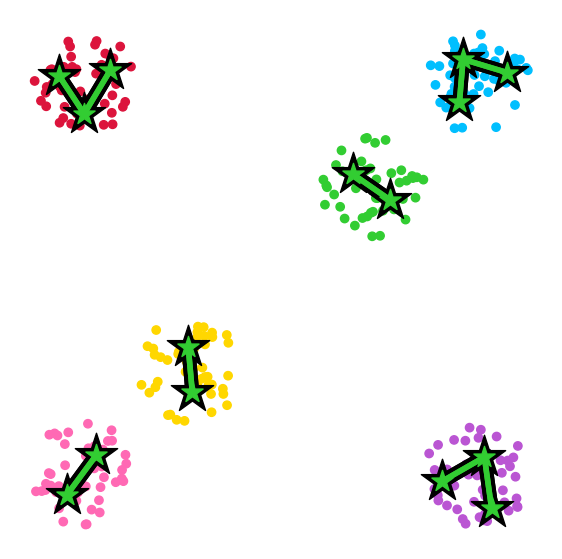}
  \end{subfigure}
  \hfill
  \begin{subfigure}[b]{0.2\linewidth}
    \includegraphics[width=\linewidth]{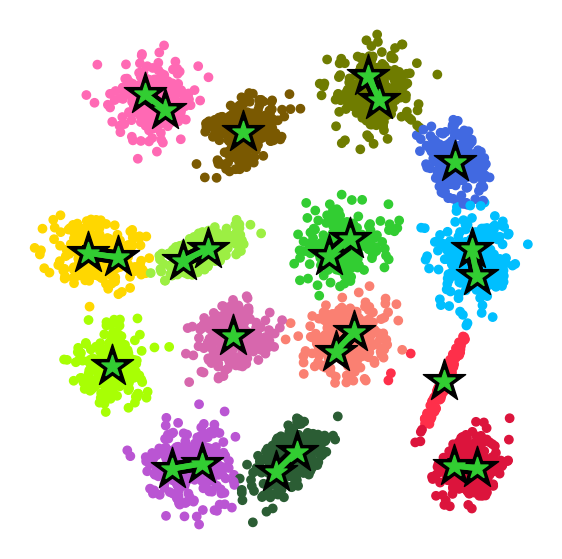}
  \end{subfigure}
  \hfill
  \begin{subfigure}[b]{0.2\linewidth}
    \includegraphics[width=\linewidth]{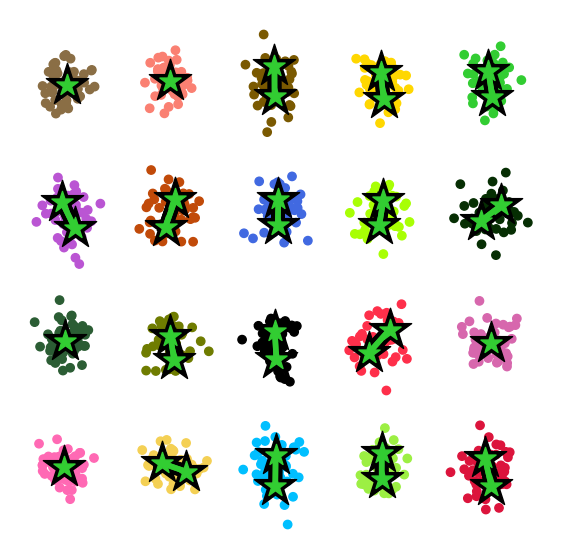}
  \end{subfigure}
  \\
  \begin{subfigure}[b]{0.2\linewidth}
    \includegraphics[width=\linewidth]{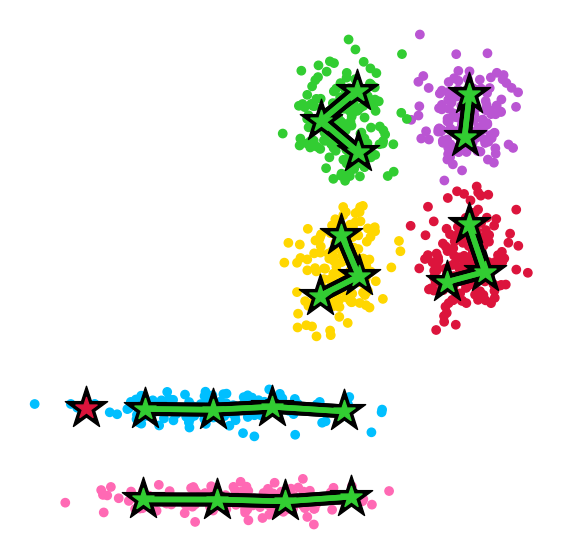}
  \end{subfigure}
  \hfill
  \begin{subfigure}[b]{0.2\linewidth}
    \includegraphics[width=\linewidth]{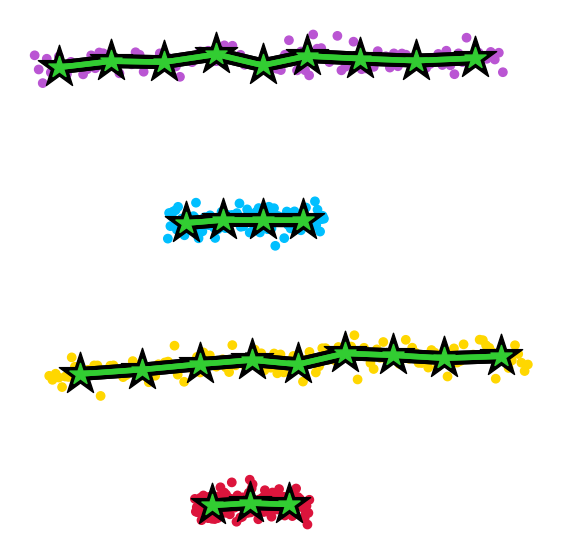}
  \end{subfigure}
  \hfill
  \begin{subfigure}[b]{0.2\linewidth}
    \includegraphics[width=\linewidth]{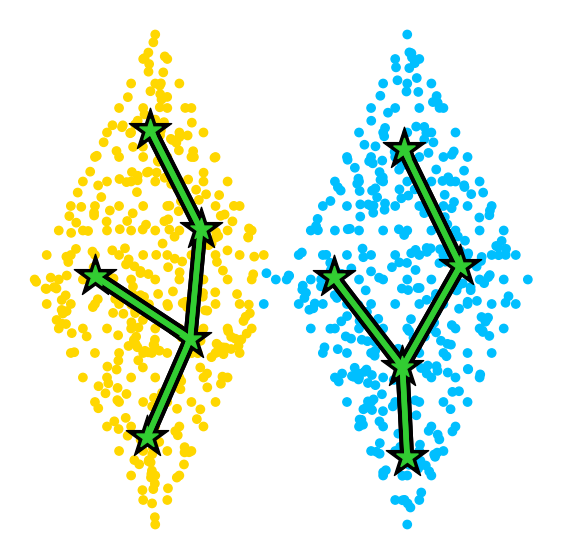}
  \end{subfigure}
  \hfill
  \begin{subfigure}[b]{0.2\linewidth}
    \includegraphics[width=\linewidth]{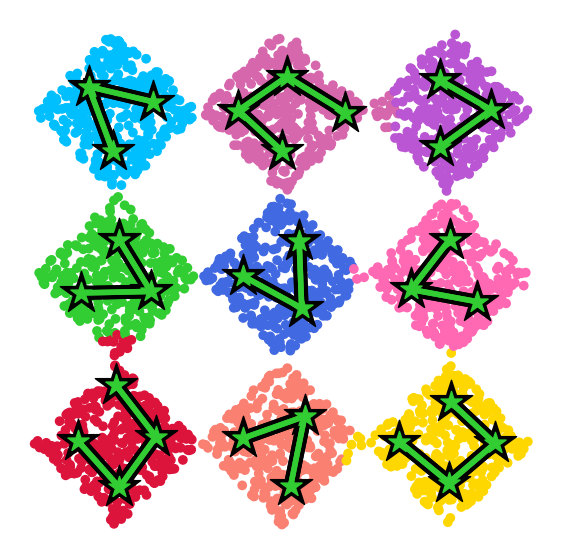}
  \end{subfigure}
  \\
  \begin{subfigure}[b]{0.2\linewidth}
    \includegraphics[width=\linewidth]{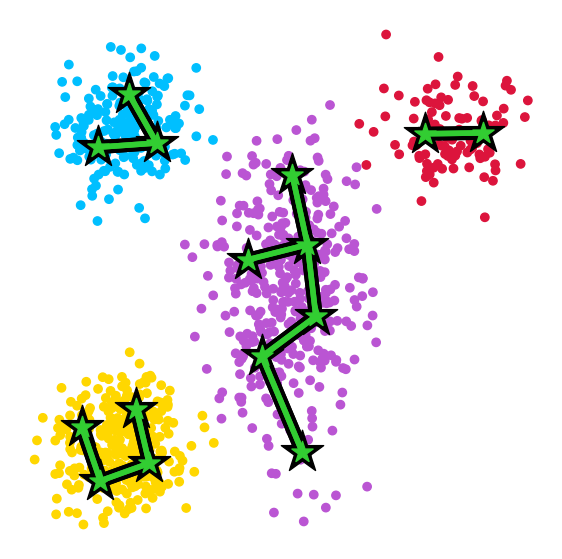}
  \end{subfigure}
  \hfill
  \begin{subfigure}[b]{0.2\linewidth}
    \includegraphics[width=\linewidth]{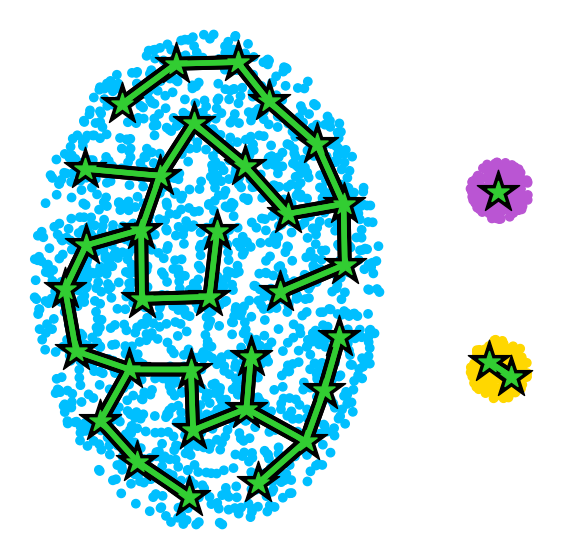}
  \end{subfigure}
  \hfill
  \begin{subfigure}[b]{0.2\linewidth}
    \includegraphics[width=\linewidth]{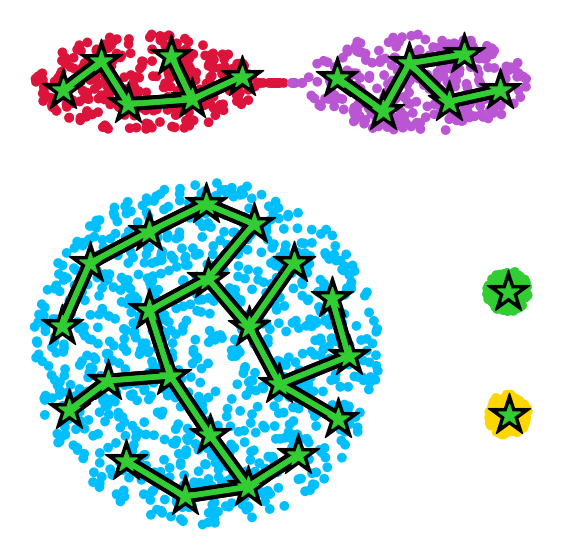}
  \end{subfigure}
  \hfill
    \begin{subfigure}[b]{0.2\linewidth}
    \includegraphics[width=\linewidth]{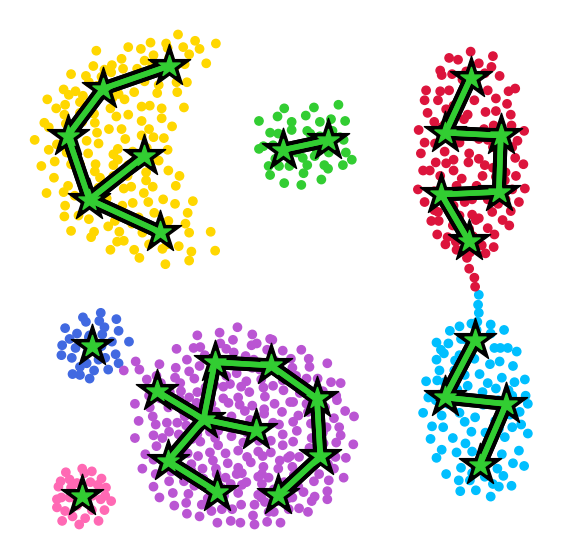}
    \end{subfigure}   
    \\
  \begin{subfigure}[b]{0.2\linewidth}
    \includegraphics[width=\linewidth]{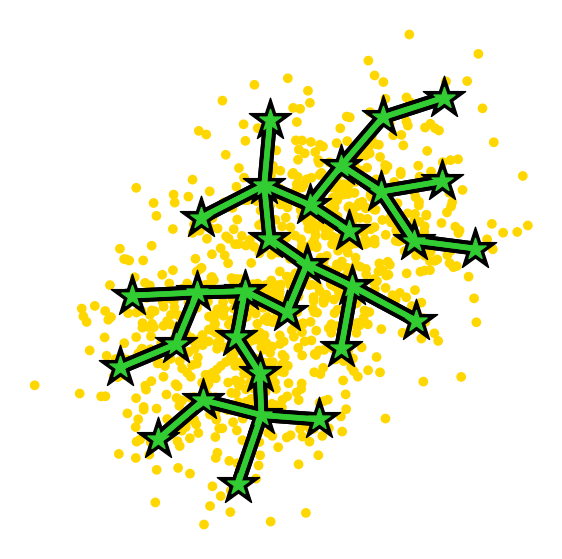}
  \end{subfigure}
  \hfill
  \begin{subfigure}[b]{0.2\linewidth}
    \includegraphics[width=\linewidth]{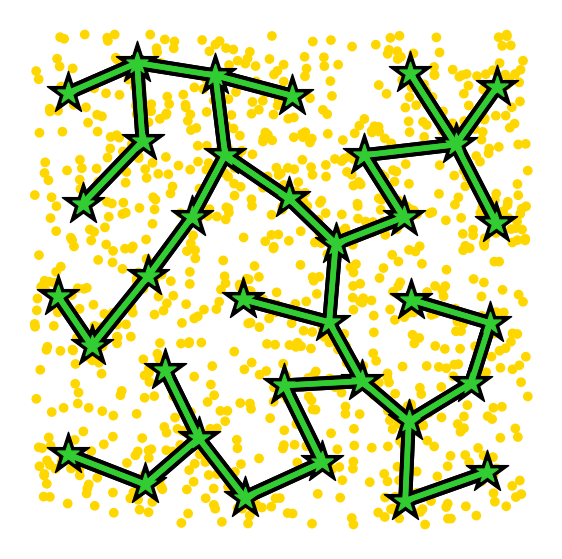}
  \end{subfigure}
  \hfill
  \begin{subfigure}[b]{0.2\linewidth}
    \includegraphics[width=\linewidth]{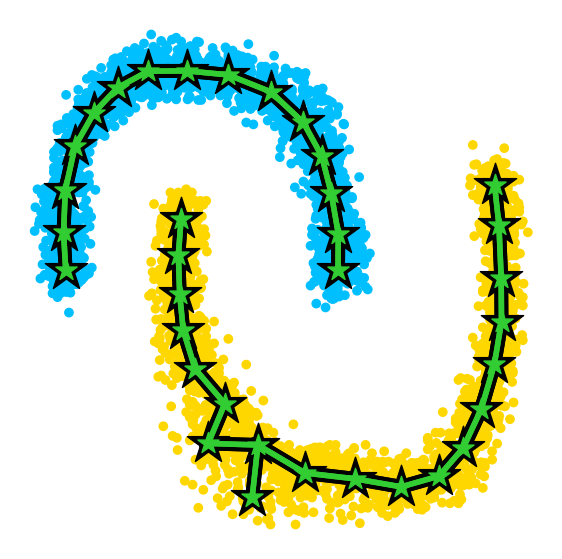}
  \end{subfigure}
  \hfill
    \begin{subfigure}[b]{0.2\linewidth}
    \includegraphics[width=\linewidth]{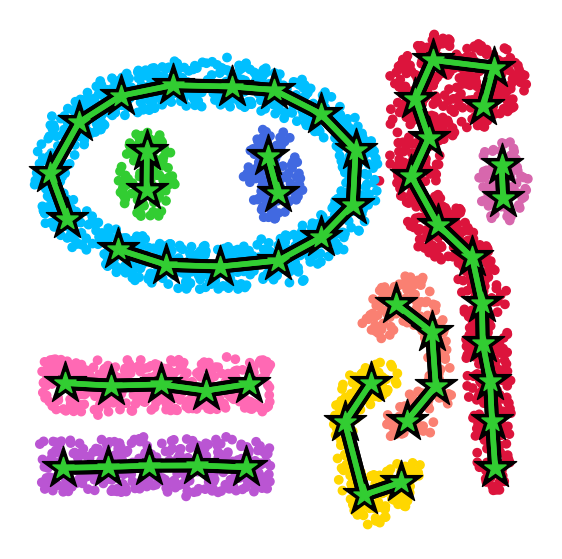}
    \end{subfigure}
    \caption{Clustering results using the \Uniforce algorithm on a large variety of $20$ synthetic datasets.}
    \label{fig:synthetic_from_literature}
\end{figure}

\subsection{Synthetic data}\label{sec:synthetic_data}
In order to provide further insight into the clustering performance of the \Uniforce algorithm, we conducted additional experiments with synthetic $2$D and $3$D datasets that have been used in the related literature\footnote{The synthetic datasets are available at: \href{https://github.com/deric/clustering-benchmark}{https://github.com/deric/clustering-benchmark}.}. Fig.\,\ref{fig:synthetic_from_literature} presents a panorama of $20$ insightful cases containing several typical and irregular shapes: Gaussian clusters, Uniform standard shapes, rings and rectangles, very elongated forms such as lines or `moons', irregular shapes, and nested clusters. In many cases, the clusters are not linearly separable and/or they are imbalanced in terms of number of datapoints and spread size. The obtained results are impressive; the locally unimodal cluster definition we propose seems versatile enough to capture the variety of data densities and shapes, and also the algorithm manages to identify meaningful clustering solutions.

Certain few cases, e.g. those in subfigures $(1,3)$ and $(3,1)$, give us the opportunity to see situations where the overclustering postprocessing decided to eliminate very small initial subclusters. The centers of those subclusters appear as red stars. Judging by these results, we can see that obtaining such small initial subclusters occurs quite rarely, and therefore the postprocessing of the overclustering is of minor importance.



\section{Conclusions}\label{sec:conclusion}
In this paper, we have proposed the \Uniforce clustering method to tackle the difficult problem of estimating the number of clusters $k$ while solving a clustering problem. Our approach is based on the novel definition of \emph{locally unimodal cluster}. The main idea is that, instead of perceiving unimodality as a property that needs to hold for the whole cluster density, we proposed to study it at a more local level, concerning subregions of the cluster density. We based our approach on the observation that unimodality may extend across pairs of neighboring subclusters when tested as a union. Such \emph{unimodal pairs} enable the aggregation of small subclusters and the bottom-up formation of larger cluster structures in a statistically sound manner. A locally unimodal cluster extends across subregions of the data density as long as there are unimodal pairs connecting them in a single connected component of the \emph{unimodality graph}. The proposed locally unimodal cluster definition is flexible as it includes typical unimodal clusters, but also encompasses non-convex and arbitrary cluster shapes. 

As part of the proposed methodology, we have developed a statistical procedure to decide on unimodal pairs of subclusters, and we built the unimodality graph in which both clustering and estimation of $k$ can be addressed through the computation of a \emph{unimodality spanning forest}. The strengths of our contribution's conceptual and algorithmic side have been  validated with extensive numerical results using several real and synthetic datasets.

Future work could focus on applying \Uniforce to various application domains to test further its versatility, and investigate the potential comparative gains of using this clustering method in existing machine learning pipelines.
Finally, we believe that the concept of local unimodality, i.e.~the fact that unimodality can be tested and validated locally, could offer a foundation for future advancements in other unsupervised learning tasks, such as density estimation, dimensionality reduction, and data visualization.

\section*{Acknowledgments}
A. Kalogeratos acknowledges support from the Industrial Data Analytics and Machine Learning Chair hosted at ENS Paris-Saclay, University Paris-Saclay.

\textbf{Code availability.}~%
An (early) version of the \Uniforce algorithm and future updates will be available in the following GitHub repository: \href{https://github.com/gvardakas/UniForCE}{https://github.com/gvardakas/UniForCE}. 
For each of the method we compared with, we used official codes made available by the authors of prior works:
The X-means, G-means, and dip-means can be found in the following site \href{https://kalogeratos.com/psite/material/dip-means/}{https://kalogeratos.com/material/dip-means/}. Pdip-means can be found in the following github repository: \href{https://github.com/Theofilos-Chamalis/Clustering-methods-based-on-statistical-testing-of-the-unimodality-of-the-data}{https://github.com/Theofilos-Chamalis/Clustering-methods-based-on-statistical-testing-of-the-unimodality-of-the-data}. A version of PG-means was provided by one of its authors. HDBSCAN and Mean Shift can be found in the scikit-learn official site: \href{https://scikit-learn.org/stable/}{https://scikit-learn.org/stable/}. The RCC implementation can be found in the following site \href{https://vladlen.info/publications/robust-continuous-clustering/}{https://vladlen.info/publications/robust-continuous-clustering/}.

{
\small
\bibliographystyle{IEEEtran}
\bibliography{references}
}

\end{document}